\documentclass{article}

\usepackage[main, final]{neurips_2026}

\usepackage[utf8]{inputenc} 
\usepackage[T1]{fontenc}    
\usepackage{hyperref}       
\usepackage{url}            
\usepackage{booktabs}       
\usepackage{amsfonts}       
\usepackage{nicefrac}       
\usepackage{microtype}      
\usepackage[table]{xcolor}        
\usepackage{graphicx}
\usepackage{wrapfig}
\usepackage{pifont}
\usepackage{amsmath}
\usepackage{amssymb}
\usepackage{mathtools}
\usepackage{amsthm}
\usepackage{algorithm}
\newcommand{\myalgorithmiccomment}[1]{\hfill\(\triangleright\)~\textcolor{gray}{#1}}
\usepackage{caption}
\definecolor{A}{RGB}{120,60,160}   
\definecolor{B}{RGB}{30,50,90}     
\definecolor{oursgray}{RGB}{235,235,235}
\definecolor{ourslavender}{RGB}{238,232,248}

\usepackage{algpseudocode} 
\usepackage{multirow}
\usepackage{makecell}    
\usepackage{siunitx}     
\usepackage[capitalize,noabbrev]{cleveref}
\usepackage{enumitem}
\usepackage{subcaption}
\theoremstyle{plain}
\newtheorem{theorem}{Theorem}[section]
\newtheorem{proposition}[theorem]{Proposition}
\newtheorem{lemma}[theorem]{Lemma}

\theoremstyle{definition}

\theoremstyle{remark}

\usepackage[disable,textsize=tiny]{todonotes}
\title{Selective Critique for Cost-Aware LLM Agents in Long-Horizon Decision Making}

\author{%
  Heewon Park$^{1}$,
  Somin Im$^{2}$,
  Minhae Kwon$^{1*}$
}

\begin{document}

\maketitle
\begingroup
\renewcommand{\thefootnote}{}
\footnotetext[0]{%
  $^{1}$Department of Electrical and Computer Engineering,
  Sungkyunkwan University, Republic of Korea.
  $^{2}$Soongsil University, Republic of Korea.
  $^{*}$Correspondence to: Minhae Kwon
  (\texttt{minhae.kwon@skku.edu}).
}
\endgroup
\setcounter{tocdepth}{1}
\addtocontents{toc}{\protect\setcounter{tocdepth}{-1}}

\begin{abstract}
Improving the reliability of large language model (LLM) agents in long-horizon decision-making remains a key challenge. When deployed as autonomous agents interacting with complex environments, early mistakes can propagate through trajectories and cause cascading failures. Recent approaches improve reliability by incorporating external critique or deliberation, but invoking these mechanisms at every step substantially increases token consumption and latency, limiting practical deployment. We propose SAG (Self-improving Agent with Gated critique), a cost-aware framework that formulates critique invocation as a step-wise decision problem during long-horizon interaction. SAG introduces a lightweight, training-free gating mechanism that estimates the utility of critique using action-level ambiguity signals—global entropy and local top-2 margin—computed over admissible actions. From a decision-theoretic perspective, this mechanism approximates the Value of Information (VoI) of critique, enabling the agent to selectively allocate expensive feedback only when its expected benefit justifies the cost. SAG further incorporates online bootstrapped self-improvement, allowing the actor to internalize critic-assisted behaviors and progressively reduce reliance on critique. Across three long-horizon interactive benchmarks and multiple backbone models, SAG substantially improves the performance-cost trade-off compared with both no-critique and always-on critique agents. On ALFWorld, SAG increases task success from $24.6\%$ to $78.3\%$ while maintaining a token budget comparable to ReAct, yielding a $3.1\times$ improvement in normalized token efficiency. Moreover, a $7$B actor with a lightweight $3$B critic achieves performance comparable to a $14$B actor without critique, showing that selective critique can recover most of the reliability benefits of deliberation while dramatically reducing inference cost. Project page: \url{https://h2w1.github.io/SAG/}
\end{abstract}

\section{Introduction}
As decision-making agents, LLMs are increasingly deployed to interact with environments and make sequential decisions, often refining their behavior through interaction experience~\cite{yao2022react,single-CoT,single-ToT,multi-RAFA,multi-debate,feedback-CER,adaplanner,eff-sequential,im2026posthoc}. Despite recent advances, long-horizon interaction remains challenging: early decision errors can propagate across steps and lead to irreversible failures~\cite{horizon_rigor,horizon_planandact,dynathink,agentpro,whydomulti,park2026multi2hierarchicalmultiagentdecisionmaking}. One route to improve reliability is to post-train agents through large-scale environment rollouts, which can produce strong task-specific policies but requires substantial training compute and infrastructure. A complementary route, and the focus of this work, is to improve the decision process at deployment time by augmenting the agent with auxiliary feedback---e.g., self-reflection, external evaluation, or reward-model based scoring---to assess candidate actions and provide corrective guidance, often improving task success~\cite{single-reflexion,critic-LAC,gou2023critic,learning-retrospex,understanding,rerest}. In this paper, we refer to such auxiliary mechanisms as \emph{critique}, and instantiate them as a \emph{critic} module queried during interaction.

Critique, however, is computationally expensive: it typically requires additional model inference and longer contexts, causing token usage, latency, and monetary cost to increase disproportionately with interaction length~\cite{glore,depo,softprompt,dellma}. This raises a central question for practical agent deployment.
\begin{center}
\emph{When should an LLM agent request a critique to balance reliability and token efficiency?}
\end{center}

\begin{wrapfigure}{rt}{0.48\textwidth}
    \centering
    \vspace{-0.5cm}
    \includegraphics[width=0.48\textwidth]{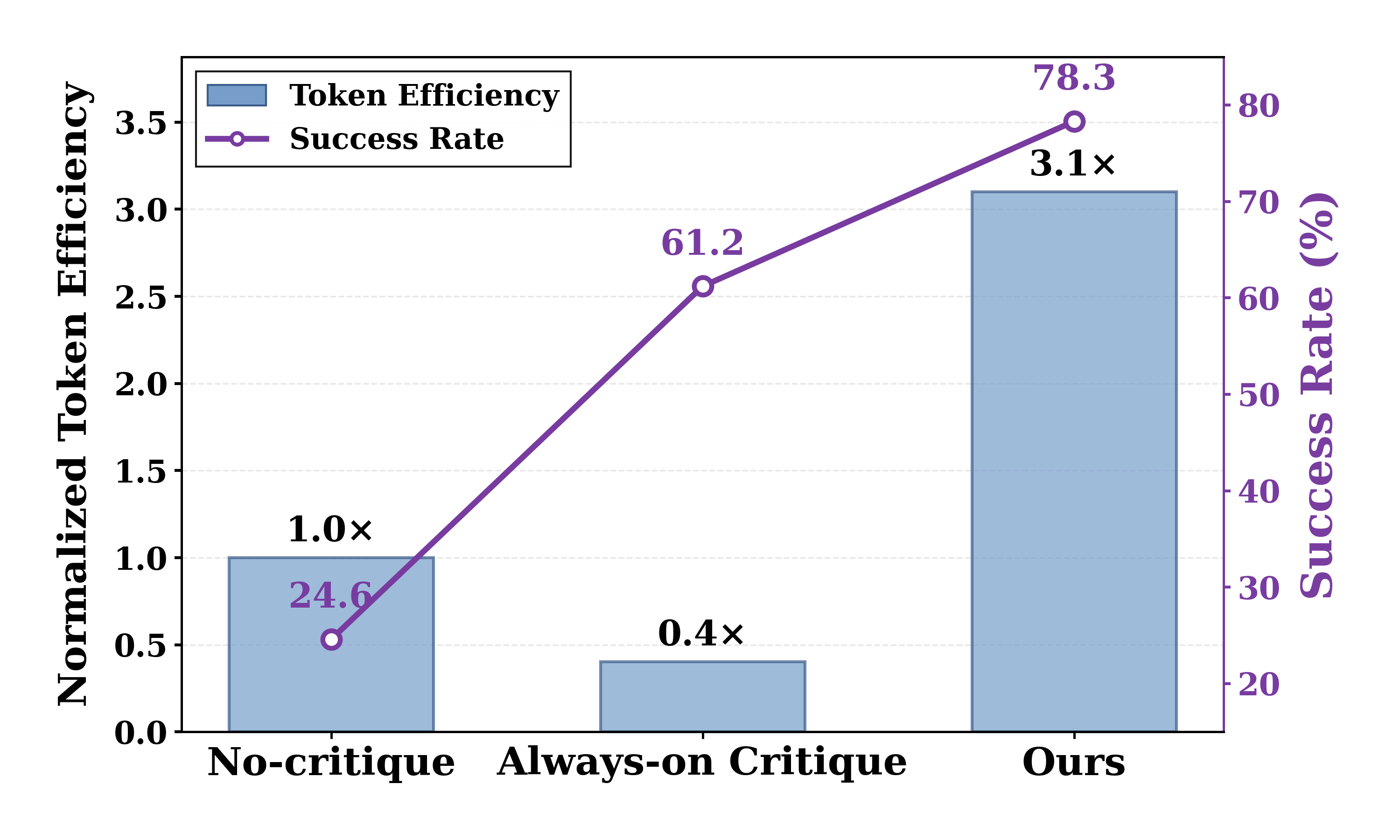}
    \vspace{-0.8cm}
    \caption{Normalized token efficiency and success rate. Token efficiency measures success rate per token usage and is normalized by ReAct (ReAct = 1.0). Always-on critique (LAC) improves success but can substantially reduce efficiency, whereas SAG achieves both high success and high token efficiency.}
    \label{fig:intro_sim}
    \vspace{-0.5cm}
\end{wrapfigure}
Figure~\ref{fig:intro_sim} highlights this performance--cost trade-off with representative instantiations (ReAct~\cite{yao2022react} and LAC~\cite{critic-LAC}). While always-on critique can improve task success, it incurs higher token usage and can reduce token efficiency, especially in long-horizon settings where critique is repeatedly invoked.

Moreover, critique is not free supervision: in realistic settings, a critic is itself an imperfect model~\cite{10684385,park2023robust} whose advice may or may not improve the next action~\cite{understanding,critic-selfplay}. More broadly, this reflects a missing design principle: many existing methods rely on fixed or heuristic critique schedules, providing limited guidance on \emph{when} critique is warranted given its cost, even as they improve \emph{how} critique is generated or applied.

\begin{figure*}[t]
    \centering
    \vspace{-0.2cm}
    \includegraphics[width=0.95\linewidth]{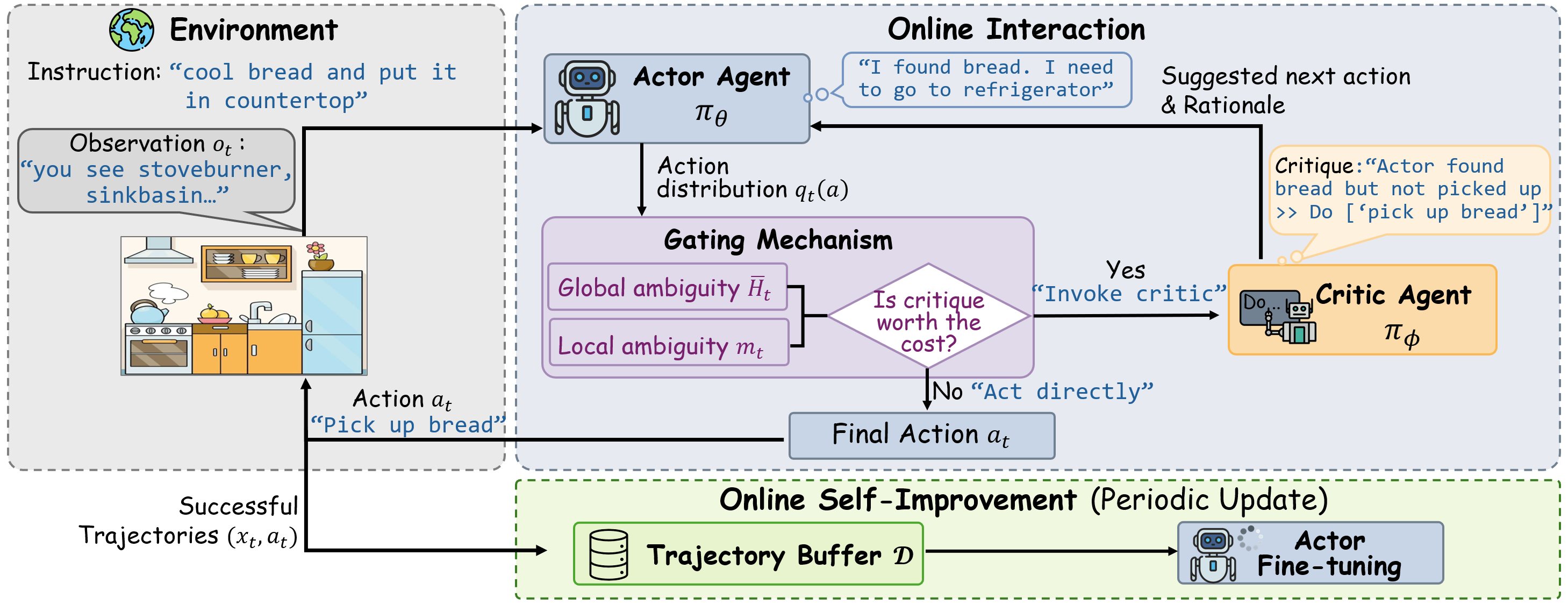}
    \vspace{-0.15cm}
    \caption{Overview of SAG. At each step, action-level ambiguity decides whether to query a critic; successful trajectories are periodically distilled into the actor via online SFT.}
    \label{fig:overview}
    \vspace{-0.5cm}
\end{figure*}

In this paper, we cast critique invocation as a \emph{cost-aware decision} rather than a fixed procedure, and introduce SAG (\textit{\textbf{S}elf-improving \textbf{A}gent with \textbf{G}ated critique}). We view critique invocation through a VoI lens: critique should be sought when its expected improvement outweighs its computational cost. While exact VoI is intractable, SAG uses the actor's action-level uncertainty as a lightweight proxy for critique value, combining global ambiguity (entropy) and local ambiguity (top-2 margin) computed over admissible action candidates. These signals trigger critique via a simple, bounded gating rule. 
Beyond selective intervention, SAG is self-improving. During online interaction, SAG stores successful trajectories and periodically fine-tunes the actor with lightweight SFT, distilling effective behaviors---including critic-assisted actions---into the policy. As the actor improves, the same VoI-motivated gate naturally reduces critique usage over time, yielding a self-evolving agent that becomes both more capable and more token-efficient. Figure~\ref{fig:overview} shows the overview of the proposed framework.


In summary, this paper makes the following contributions. 1) \textbf{SAG: cost-aware gated critique for LLM agents.} 
We propose SAG, a framework that formulates critic invocation as a step-wise decision problem during long-horizon interaction. By selectively allocating critique only when it is expected to improve decision quality, SAG enables agents to balance reliability with token efficiency, directly addressing the inference cost and latency bottlenecks of practical LLM agent deployment. 
2) \textbf{Lightweight ambiguity-based gating.} 
SAG introduces a simple, training-free gating mechanism that combines global ambiguity (normalized entropy) and local ambiguity (top-2 margin) over admissible action candidates. Despite its simplicity, this mechanism provides an effective and scalable strategy for adaptive critique allocation without additional models or training. 
3) \textbf{A principled VoI perspective on critique.} 
We interpret critique invocation through a decision-theoretic lens and show that actor-side ambiguity signals serve as practical proxies for the expected utility of critique, providing a theoretically grounded basis for cost-aware reasoning allocation in LLM agents. 
4) \textbf{Critique-driven online self-improvement.} 
SAG leverages selectively refined trajectories as supervision signals for online bootstrapped improvement. As critic-assisted behaviors are distilled into the actor, the agent progressively improves decision quality while naturally reducing reliance on critique over time. 
5) \textbf{Comprehensive empirical validation of success–cost efficiency.}
Across three long-horizon interactive environments, multiple backbone models, and extensive ablations, scaling studies, and matched self-improvement controls, SAG consistently improves the success–cost Pareto frontier over both no-critique and always-on critique baselines. On ALFWorld, SAG increases success from $24.6\%$ to $78.3\%$ while maintaining a token budget comparable to ReAct, yielding a $3.1\times$ improvement in normalized token efficiency. SAG further shows larger gains on harder tasks and narrows the gap between smaller and larger models, with a 7B actor and 3B critic matching the performance of a 14B actor without critique.

\section{Related Work}
\paragraph{LLM agents in interactive environments}
LLMs are increasingly used as agents that reason and act in interactive environments. Existing work improves such agents along two broad directions. The first improves an available agent during execution through prompting, deliberation, feedback, or lightweight adaptation~\cite{yao2022react,multi-debate,single-reflexion,self-refine,fed-ade,10.1145/3746252.3760830,im2025personalized,11143204,joe2024finetuning}. ReAct~\cite{yao2022react} interleaves reasoning, actions, and observations for multi-step decision making, and subsequent work improves robustness with self-reflection~\cite{single-reflexion, self-refine}, debate~\cite{multi-debate, debate2, debate3}, confidence-based aggregation~\cite{feedback-CER, conf-cisc}, and trajectory reuse or bootstrapping~\cite{ICL, CGI, se-agent}. Inference under partial or imperfect observations has also been studied~\cite{
pmlr-v306-wang26ab,
10.1145/3746252.3761304,
10.1145/3711875.3734549, joe2024finetuning,
e84a08272b4649948199613b5988465c,
3fe73e36f06442c68a9362beaffa8171,
kim2025cluvar,
KWON2025109919,
wu2020rational,
NEURIPS2020_5a01f059}.
The second improves the policy before deployment~\cite{eo2024offline,im2026posthoc} through preference optimization, environment rollouts, or post-training reinforcement learning (RL)~\cite{dpo, ppo, gigpo, grpo}, often investing substantial task-specific compute before the agent is deployed. In this paper, we target the former deployment-time regime: given an available actor and a non-oracle critic, we ask when critique is worth its cost during interaction.

\vspace{-0.2cm}
\paragraph{Critique and cost-aware improvement}
Explicit critique and feedback can correct agent behavior and support learning from experience~\cite{critic-LAC, learning-retrospex, proactive, CGI, webevolver,eom2023autonomous,eom2024selective}. Prior critic-guided methods use continuous feedback, self-asking, or trajectory-level evaluation to improve agent decisions~\cite{critic-LAC, CGI, prospector, learning-retrospex}, while self-improvement methods train on successful sampled trajectories~\cite{rft}. However, these approaches often rely on frequent critique, repeated sampling, or fixed schedules, incurring overhead from extra model calls, longer contexts, or multiple candidate evaluations~\cite{critic-selfplay, critic-selff, budget, costaware}. Although adaptive-computation methods allocate extra inference using confidence or uncertainty~\cite{agile}, critique in long-horizon agents is still often invoked by fixed heuristics~\cite{critic-copper, yuan2025reinforce, prospector,hybirdllm}. In contrast, our work frames critique invocation as a cost-aware step-wise decision problem, selectively querying a non-oracle critic using lightweight ambiguity signals that we later justify as proxies for the VoI.

\vspace{-0.2cm}
\section{Gated Critic-guided Self-improving Agent}
\label{sec:method}
\vspace{-0.2cm}
We propose SAG, an efficient LLM agent framework that selectively invokes critique via a cost-sensitive gate and internalizes successful critic-guided behavior through online learning. Algorithm~\ref{alg:sag} summarizes SAG, and Section~\ref{sec:theory} motivates the ambiguity gate through a VoI analysis. The notations used in this paper are summarized in Appendix~\ref{app:notation}.

\vspace{-0.2cm}
\subsection{Problem Formulation}
\label{sec:problem}

We consider an interactive environment in which an LLM-based actor performs a sequence of language actions to solve a task. At time step $t$, the actor receives an observation $o_t$ and forms a decision context $x_t$ that contains the task instruction, the current observation, and a truncated interaction history. The environment provides an admissible action set $\mathcal{A}_t$ containing valid actions at step $t$.

Let $\pi_\theta$ denote the actor policy. The actor first produces a reasoning state $r_t$ from $x_t$ and scores each admissible action $a\in\mathcal{A}_t$ by its length-normalized token log-likelihood, i.e., 
$\ell_t(a)=\frac{1}{|a|}\sum_{j=1}^{|a|}\log p_\theta(a_j\mid x_t,r_t,a_{<j}).
$
We define the actor-induced distribution over admissible actions as
$q_t(a)=
\frac{\exp(\ell_t(a))}
{\sum_{a'\in\mathcal{A}_t}\exp(\ell_t(a'))}$ for $a\in\mathcal{A}_t$, 
and compute ambiguity over this action-level distribution.

SAG additionally introduces a binary critic-invocation decision $g_t\in\{0,1\}$. $g_t=1$ queries a critic before finalizing an action, while $g_t=0$ executes the actor decision immediately. Querying the critic incurs an additional cost $c>0$. From a decision-theoretic viewpoint, critic invocation can be interpreted as a VoI choice: critique is invoked only when its expected improvement exceeds the cost. We formalize this principle in Section~\ref{sec:theory} and in practice approximate it using ambiguity signals derived from the actor.

\begin{wrapfigure}{r}{0.55\textwidth}
    \begin{minipage}{0.55\textwidth}
    \setlength{\textfloatsep}{8pt}
        \vspace{-0.85cm}
        \footnotesize
        \begin{algorithm}[H]
            \caption{SAG}
            \label{alg:sag}
            \begin{algorithmic}[1]
            \Require Actor policy $\pi_\theta$, critic policy $\pi_\phi$, environment $\mathcal{E}$, thresholds $(\tau,\delta)$

            \State Observe context $x_t$ and admissible actions $\mathcal{A}_t$
            \State Generate reasoning state $r_t$ from $x_t$
            \State Compute actor distribution $q_t(a)$ over $\mathcal{A}_t$

            \State \textcolor{A}{\textbf{Compute actor ambiguity}}
            \State $\bar H_t \gets$ normalized entropy of $q_t$ \myalgorithmiccomment{global ambiguity}
            \State $m_t \gets$ log-prob margin of $q_t$ \myalgorithmiccomment{local ambiguity}

            \State \textcolor{A}{\textbf{Gated critic decision}}
            \State $S_t \gets \bar H_t + \mathbb{I}[m_t < \delta]$
            \State $g_t \gets \mathbb{I}[S_t > \tau]$ \myalgorithmiccomment{critic gate}

            \If{$g_t = 1$}
                \State Query critic $\pi_\phi$ with $(x_t,r_t,\mathcal{A}_t)$
                \State $a_t \gets$ critic-suggested action
            \Else
                \State $a_t \gets \arg\max_{a \in \mathcal{A}_t} q_t(a)$
            \EndIf

            \State Execute $a_t$ in environment

            \State \textcolor{B}{\textbf{Online self-improvement (periodic)}}
            \If{episode succeeds}
                \State Store successful trajectories in buffer $\mathcal{D}$
            \EndIf
            \If{update time}
                \State Update actor via SFT on $\mathcal{D}$
            \EndIf
            \end{algorithmic}
        \end{algorithm}
        \vspace{-0.6cm}
    \end{minipage}
    \vspace{-0.2cm}
\end{wrapfigure}

\subsection{Ambiguity Signals from the Actor}
SAG estimates when feedback from the critic is likely to be beneficial using ambiguity signals derived from the actor-induced distribution $q_t$. These signals capture complementary aspects of decision ambiguity.

\vspace{-0.2cm}
\paragraph{Global ambiguity}
We measure global ambiguity via normalized entropy.
\begin{equation}
\label{eq:norm_entropy}
\bar H_t = -\frac{1}{\log |\mathcal{A}_t|} \sum_{a\in\mathcal{A}_t} q_t(a)\log q_t(a)
\end{equation}
This is defined for $|\mathcal{A}_t|\ge 2$ and set to $0$ otherwise. High entropy indicates many-way ambiguity among admissible actions.

\vspace{-0.2cm}
\paragraph{Local ambiguity}
To capture ambiguity near a decision boundary, let $a_1$ and $a_2$ denote the top-1 and top-2 actions under $q_t$. We define the log-probability margin as
\begin{equation}
\label{eq:margin}
m_t = \log q_t(a_1) - \log q_t(a_2).
\end{equation}
A small margin indicates that two competing actions are nearly indistinguishable, making the decision sensitive to additional information. When $|\mathcal{A}_t|<2$, we set $m_t=\infty$, so the local ambiguity indicator is inactive.

\vspace{-0.2cm}
\subsection{Gated Critic Invocation}
\vspace{-0.2cm}
SAG combines global and local ambiguity into a bounded gating score with margin threshold $\delta$. 
\begin{equation}
\label{eq:gating_score}
S_t \;=\; \bar H_t + \mathbb{I}[m_t < \delta], \quad S_t\in[0,2]
\end{equation}
We use a binary margin indicator rather than a continuous function to detect near-boundary decisions, since the goal of the margin term is not fine-grained calibration but robust identification of high-risk local ambiguities where small additional information can flip the decision. The critic is invoked when 
\begin{equation}
\label{eq:gating_rule}
g_t \;=\; \mathbb{I}[S_t > \tau],
\end{equation}
where $\tau\in(0,2)$ is a cost-calibrated threshold. Intuitively, $\tau$ is chosen as a monotone function of the critic cost $c$: higher costs result in more conservative critic usage.

When $g_t=1$, SAG queries a non-oracle critic $\pi_\phi$ with the decision context, reasoning state, and admissible actions. The critic has no access to ground-truth actions or environment rewards; it returns a diagnostic rationale and a suggested action, which may itself be incorrect. If the critic output cannot be parsed as an admissible action, SAG falls back to the actor's top-1 action under $q_t$. In our main experiments, the actor and critic use the same backbone with different role prompts, so gains are not attributed to a stronger critic model but to selective use of critique.

\vspace{-0.2cm}
\subsection{Online Self-Improvement via Bootstrapped Supervision}
\label{sec:self_improve}
\vspace{-0.2cm}
SAG continuously improves the actor by internalizing successful behaviors observed during online interaction. Episodes are processed sequentially; after a successful episode, we collect context--action pairs $(x_t,a_t)$ into a dataset $\mathcal{D}$, where $a_t$ is the executed admissible action, either selected by the actor or suggested by the critic when critique is invoked. The critic is kept fixed, and every $K$ successful episodes, only the actor is periodically updated using action-level SFT:
$\mathcal{L}_{\mathrm{SFT}}(\theta)
= \mathbb{E}_{(x,a)\sim \mathcal{D}}\big[ -\log \pi_\theta(a\mid x) \big]$.
The effect of $K$ is analyzed in Appendix~\ref{app:update_sensitivity}. Since the target is action-only, SAG does not train on mismatched actor reasoning traces paired with critic-corrected actions.

Following prior bootstrapping-based self-training~\cite{zelikman2022star}, we perform each update by fine-tuning from the same base model whenever new trajectories are accumulated, rather than continuing from the previous updated actor. This avoids compounding errors and preserves distributional stability. This resembles rejection fine-tuning in using successful trajectories, but SAG collects them online under a gated critique policy, allowing critic-assisted corrections to become action-level supervision while explicitly controlling critique cost.
\vspace{-0.2cm}
\section{Theoretical Analysis}
\label{sec:theory}
\vspace{-0.2cm}
We provide a decision-theoretic justification for SAG. Our goal is to show that (i) critic invocation admits a principled VoI interpretation, and (ii) the actor-side ambiguity signals used in SAG serve as lightweight proxies for VoI.
\vspace{-0.3cm}

\subsection{Critic as a VoI Decision}
\label{sec:theory_voi}
\vspace{-0.2cm}
At each step $t$, let $z_t=(x_t,r_t)$ denote the actor's current decision state after reasoning. SAG chooses $g_t\in\{0,1\}$. Invoking the critic ($g_t=1$) incurs an additional cost $c>0$. Let $\Delta_t$ denote the step-level improvement attributable to invoking critique under decision state $z_t$ (e.g., an increase in success probability). We define the VoI of critique as
\begin{equation}
\label{eq:voi_def}
\mathrm{VoI}(z_t):=\mathbb{E}[\Delta_t\mid z_t].
\end{equation}

\begin{lemma}[VoI thresholding for critic invocation]
\label{lem:voi_gate}
For the binary decision of whether to invoke the critic at step $t$, the Bayes-optimal rule is
\begin{equation}
\label{eq:voi_opt_gate}
g_t^\star \;=\; \mathbb{I}\!\left[\mathrm{VoI}(z_t) > c\right]
\;=\;
\mathbb{I}\!\left[\mathbb{E}[\Delta_t \mid z_t] > c\right].
\end{equation}
\end{lemma}

\vspace{-0.2cm}
Proof for Lemma~\ref{lem:voi_gate} is provided in Appendix~\ref{app:voi_gate}. In practice, $\mathrm{VoI}(z_t)$ is intractable to compute, which motivates using lightweight proxy signals derived from the actor.
\vspace{-0.2cm}

\subsection{Ambiguity as a Proxy for VoI}
\label{sec:theory_proxy}
\vspace{-0.2cm}
We next describe why the actor-side ambiguity signals in SAG provide lightweight proxies for the VoI term $\mathbb{E}[\Delta_t\mid z_t]$ in Eq.~\eqref{eq:voi_opt_gate}. The key observation is that critique is valuable mainly when the actor is likely to be wrong. In particular, if critique corrects the actor with probability $\rho$ when the actor is wrong, and a successful correction yields average improvement $\gamma$, then the expected step-level gain can be approximated by
\begin{equation}
\label{eq:voi_error_risk_compact}
\mathbb{E}[\Delta_t\mid z_t]\approx\rho\,\gamma\Pr(\mathrm{err}_t\mid z_t),
\end{equation}
where $\mathrm{err}_t$ denotes the event that the actor's immediate decision at decision state $z_t$ is incorrect. 
Combined with Lemma~\ref{lem:voi_gate}, this approximation implies that critique should be invoked at high-risk steps. In particular, the VoI condition $\mathbb{E}[\Delta_t \mid z_t] > c$ is equivalent to $\Pr(\mathrm{err}_t \mid z_t) > \tfrac{c}{\rho\gamma}$, where the same critic cost $c$ is expressed in the error-risk space by scaling with the correction reliability $\rho$ and average gain $\gamma$.
Since $\Pr(\mathrm{err}_t\mid z_t)$ is not directly observable, we next relate it to ambiguity properties of $q_t$.

\textbf{Global ambiguity}
Let $a_1=\arg\max_{a\in\mathcal{A}_t} q_t(a)$ and $q_{\max}=q_t(a_1)$. We use normalized entropy $\bar H_t$ (Eq.~\ref{eq:norm_entropy}) to capture global ambiguity across candidate actions.

\begin{proposition}[Entropy controls greedy error risk]
\label{prop:entropy_err}
Interpreting $q_t$ as the actor's belief over which admissible action is correct at decision state $z_t$, the step-level error risk satisfies
\begin{equation}
\label{eq:err_eq_1minusqmax}
\Pr(\mathrm{err}_t\mid z_t)\;=\;1-q_{\max}.
\end{equation}
Moreover, with entropy $H(q_t)=-\sum_{a\in\mathcal{A}_t} q_t(a)\log q_t(a)$, we have $q_{\max}\ge \exp(-H(q_t))$. Thus, we get
\begin{equation}
\label{eq:entropy_err_bound}
\Pr(\mathrm{err}_t\mid z_t) \le 1-\exp\left(-H(q_t)\right) =1-|\mathcal{A}_t|^{-\bar H_t}.
\end{equation}
\end{proposition}

Proof for Proposition~\ref{prop:entropy_err} is provided in Appendix~\ref{app:entropy_bounds}. Proposition~\ref{prop:entropy_err} formalizes that higher entropy $\bar H_t$ rules out the presence of a single dominant action, necessarily increasing an upper bound on the error risk of greedy execution. This corresponds to a regime in which probability mass is spread across many admissible actions, rather than concentrated on a clear best choice. We refer to this regime as global ambiguity, since ambiguity is distributed broadly over the action space. In such decision states, additional information from critique has greater potential value, motivating the use of entropy as a proxy for global decision risk.

\textbf{Local ambiguity} 
Global ambiguity can under-penalize near-ties between the top two actions (e.g., when $q_t$ concentrates on two candidates).
We therefore also use the margin $m_t$ (Eq.~\ref{eq:margin}) to capture \emph{near-boundary} uncertainty.

\begin{proposition}[Margin controls top-2 error risk]
\label{prop:margin_top2}
Assume the correct action lies in $\{a_1,a_2\}$. Then selecting $a_1$ has a conditional error probability as follows.
\begin{equation}
\label{eq:margin_err}
\Pr(\mathrm{err}\mid z_t,\mathrm{top\text{-}2})= \frac{q_t(a_2)}{q_t(a_1)+q_t(a_2)}= \frac{1}{1+e^{m_t}}\le e^{-m_t}
\end{equation}
\end{proposition}

Proof for Proposition~\ref{prop:margin_top2} is provided in Appendix~\ref{app:margin_risk}. Proposition~\ref{prop:margin_top2} formalizes that small margins correspond to large error risk, making the decision sensitive to small amounts of additional information. Such near-boundary decisions are precisely where critique is most likely to change the outcome, motivating the use of the margin as a proxy for local decision risk.

Therefore, decision states with higher error risk admit larger VoI. Proposition~\ref{prop:entropy_err} and Proposition~\ref{prop:margin_top2} show that (i) high entropy rules out a dominant action (broad ambiguity), and (ii) small margin implies large local error risk. This motivates using $\bar H_t$ and $m_t$ as lightweight, computable proxies for VoI in the gating score (Eqs.~\eqref{eq:gating_score} and~\eqref{eq:gating_rule}).

\vspace{-0.2cm}
\section{Experimental Setting}
\label{experimental_setting}
\vspace{-0.2cm}
In this section, we describe simulation settings for Section~\ref{sec:main_results}. Additional simulation details are provided in Appendix~\ref{app:setting}.

\textbf{Interactive decision-making benchmarks} 
We evaluate SAG on three complementary long-horizon benchmarks: \textbf{ALFWorld}~\cite{shridhar2020alfworld}, which requires grounded action under partial observability; \textbf{BabyAI-Text}~\cite{babyai}, which tests compositional instruction following and planning; and \textbf{WebShop}~\cite{yao2022webshop}, which involves noisy open-domain search and exploration. Together, they cover state tracking, multi-step reasoning, and cost-aware decision making. Details are in Appendix~\ref{app:benchmarks}.


\textbf{Backbone Models and Training}
We evaluate all methods using three instruction-tuned LLM backbones: Qwen2.5-7B-Instruct~\cite{qwen2.5}, Llama-3.1-8B-Instruct~\cite{llama3modelcard}, and Mistral-7B-Instruct~\cite{mistral}. For SAG, the actor and critic use the same backbone and model size unless otherwise stated, differing only in their role prompts. Actor--critic capacity scaling is analyzed with Qwen2.5-Instruct 3B/7B/14B pairings and extended to 32B in Appendix~\ref{app:extended_scaling}. SAG performs online self-improvement via action-level SFT: context--action pairs from successful episodes are collected, and the actor is updated whenever 10 successful trajectories are accumulated. Unless otherwise stated, all baselines share the same backbone models and inference settings to ensure fair comparison. The online self-improvement protocol and hyperparameters are provided in Appendix~\ref{app:online_protocol} and~\ref{app:hyperparameter}.

\textbf{Baselines and comparison methods} 
We compare SAG against representative LLM agent frameworks, grouped into three categories. We focus on inference-time and lightweight-adaptation baselines; the scope relative to RL-based agent post-training is discussed in Appendix~\ref{app:rl_scope}.
\begin{itemize}[leftmargin=*, itemsep=0em, topsep=0em]
    \item \textbf{Basic single-agent baseline} (ReAct~\cite{yao2022react})
    which interleaves reasoning and environment actions, serving as a lightweight baseline that does not invoke a critic during execution.
    \item \textbf{Deliberation methods} (Reflexion~\cite{single-reflexion}, Debate~\cite{multi-debate}, CER~\cite{feedback-CER}, Traj-Boost~\cite{ICL}) which improve robustness by increasing computation through multi-pass deliberation or experience reuse (e.g., self-reflection across trials, multi-agent discussion, confidence-based selection/aggregation, or trajectory reuse). However, they do not model critique invocation as a cost-sensitive, step-wise decision during execution.
    \item \textbf{Always-on critic methods} (Retrospex~\cite{learning-retrospex}, LAC~\cite{critic-LAC})
    which utilize a critic throughout execution by re-scoring or querying at essentially every step. While effective for success, their persistent critic usage can substantially increase inference cost and reduce token efficiency.
        \item \textbf{High-capacity reference} (ReAct using GPT-4.1~\cite{achiam2023gpt})
    which provides a reference point for how a much stronger model performs without critique.
\end{itemize}

\begin{wrapfigure}{r}{0.45\textwidth}
    \centering
    \vspace{-0.4cm}
    \includegraphics[width=0.45\textwidth]{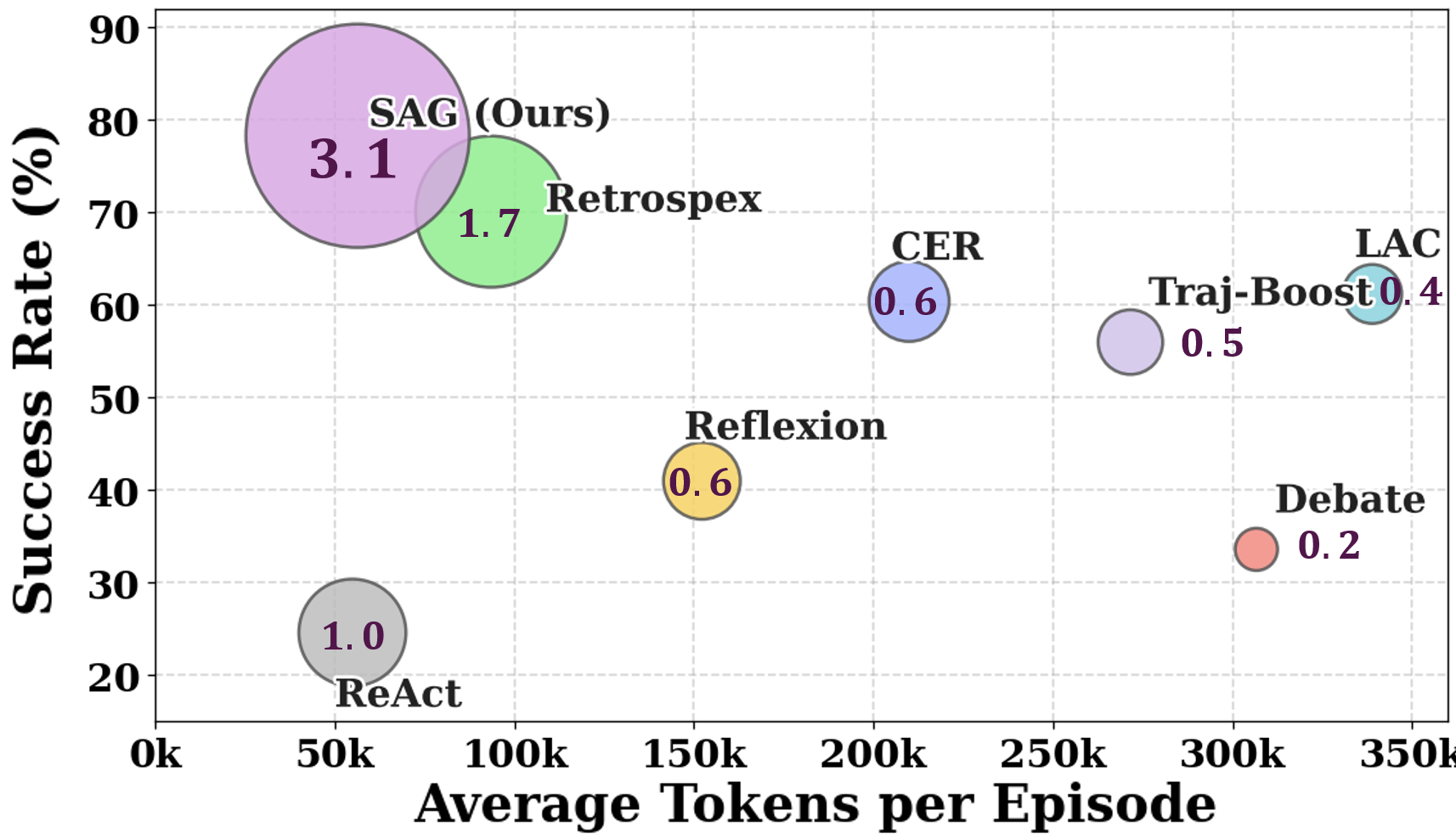}
    \caption{Performance--cost comparison. Bubble area indicates normalized token efficiency (ReAct = $1.0$).}
    \label{fig:alfworld_tradeoff_qwen}
    \vspace{-0.3cm}
\end{wrapfigure}
\textbf{Evaluation protocol and metrics} 
We evaluate all methods on standard test sets (ALFWorld: 134; BabyAI-Text: 100; WebShop: 200 instances) using a sequential online-evaluation protocol. For SAG, each episode is evaluated once; successful past trajectories may update the actor before later episodes, but no episode is re-evaluated after being used for self-improvement. We report success rate, average steps per successful episode, average tokens per episode, and normalized token efficiency, defined as success rate per token usage and normalized by ReAct for each benchmark and backbone. All methods use pass@1, except Reflexion in its native pass@3 setting. Detailed analyses and ablations in Section~\ref{sec:main_results} are conducted on ALFWorld with Qwen2.5-7B-Instruct, which serves as our primary setting.

\begin{table}[t]
\vspace{-1cm}
\centering
\scriptsize
\setlength{\tabcolsep}{3.5pt}
\renewcommand{\arraystretch}{1.2}
\caption{Performance comparison on three benchmarks with three backbones.
\textbf{Succ}: success rate (\%).
\textbf{\#Horizon}: average steps per successful episode.
\textbf{\#Token}: average tokens per episode (k).
\textbf{Eff}: normalized token efficiency, computed as $\text{Succ}/\text{\#Token}$ and normalized so that ReAct = 1.0 for each backbone and benchmark.}
\label{tab:main_results}

\resizebox{\linewidth}{!}{%
\begin{tabular}{lllc cccccccccccc}
\toprule
\textbf{Model} & \textbf{Type} & \textbf{Method}
& \multicolumn{4}{c}{\textbf{ALFWorld}} 
& \multicolumn{4}{c}{\textbf{BabyAI}} 
& \multicolumn{4}{c}{\textbf{WebShop}} \\
\cmidrule(lr){4-7} \cmidrule(lr){8-11} \cmidrule(lr){12-15}
& & 
& \textbf{Succ}$\uparrow$& \textbf{\#Horizon}$\downarrow$& \textbf{\#Token}$\downarrow$& \textbf{Eff}$\uparrow$
& \textbf{Succ}$\uparrow$& \textbf{\#Horizon}$\downarrow$& \textbf{\#Token}$\downarrow$& \textbf{Eff}$\uparrow$
& \textbf{Succ}$\uparrow$& \textbf{\#Horizon}$\downarrow$& \textbf{\#Token}$\downarrow$& \textbf{Eff}$\uparrow$\\
\midrule

\multirow{1}{*}{GPT-4.1} 
& Basic & ReAct 
& $80.6$ & $11.9$ & $32$ & $1.00$
& $74.0$ & $15.2$ & $27$ & $1.00$
& $33.5$ & $7.9$ & $10$ & $1.00$ \\
\midrule

\multirow{8}{*}{\rotatebox{90}{\shortstack{\textbf{Qwen2.5-}\\\textbf{7B-Instruct}}}}
& Basic        & ReAct      
& $24.6$ & $15.8$ & $\mathbf{55}$  & $1.00$ & $36.0$ & $7.4$ & $55$ & $1.00$ & $11.0$ & $6.3$ & $12$ & $1.00$ \\
& Deliberation & Reflexion  
& $40.3$ & $16.3$ & $150$ & $0.60$ & $41.0$ & $\mathbf{7.0}$ & $162$ &$ 0.39$ & $11.5$ & $6.8$ & $15$ & $0.81$ \\

& Deliberation    & Debate     
& $31.3$ & $14.3$ & $389$ & $0.18$ & $52.0$  &$ 8.9$ & $441$ &$ 0.18$ & $20.0$ & $\mathbf{4.7}$ &$ 90$ & $0.24$ \\

& Deliberation  & CER        
& $60.5$ &$ 16.4$ & $210$ &$ 0.64$ &$ 45.0$ &$ 8.0$ & $239$ & $0.29$ & $16.0$ & $5.1$ & $11$ & $1.50$ \\

& Deliberation        & Traj-Boost 
& $56.0$ & $12.9$ & $272$ & $0.46$ & $44.0$ & $10.0 $& $85$ & $0.80$ & $24.5$ & $5.1$ & $12$ & $2.19$ \\

& Critic   & Retrospex  
& $70.2$ & $\mathbf{10.5}$ & $94$ & $1.67$ & $42.0$ & $9.5$ & $\mathbf{50}$ & $1.30$ & $20.0$ & $5.9$ & $\mathbf{6}$ & $3.58$ \\

& Critic     & LAC        
& $61.2$ & $14.2$ & $339$ & $0.40$ & $47.0$ & $10.5$ & $436$ & $0.17$ & $25.0$ & $6.5$ & $82$ & $0.32$ \\
& \cellcolor{ourslavender}\textbf{Critic }
& \cellcolor{ourslavender}\textbf{Ours (SAG)}
& \cellcolor{ourslavender}$\mathbf{78.3}$
& \cellcolor{ourslavender}$11.4$
& \cellcolor{ourslavender}$56$
& \cellcolor{ourslavender}$\mathbf{3.10}$
& \cellcolor{ourslavender}$\mathbf{70.0}$
& \cellcolor{ourslavender}$8.1$
& \cellcolor{ourslavender}$78$
& \cellcolor{ourslavender}$\mathbf{1.38}$
& \cellcolor{ourslavender}$\mathbf{37.0}$
& \cellcolor{ourslavender}$5.8$
& \cellcolor{ourslavender}$9$
& \cellcolor{ourslavender}$\mathbf{4.55}$ \\

\midrule
\multirow{8}{*}{\rotatebox{90}{\shortstack{\textbf{Llama-3.1-}\\\textbf{8B-Instruct}}}}
&  Basic        & ReAct      
& $51.5$ & $15.1$ & $\mathbf{45}$ &$ 1.00 $& $32.0$ & $\mathbf{6.6}$ & $\mathbf{55}$ & $1.00$ & $26.5 $& $7.1$ & $9$ & $1.00$ \\
& Deliberation  & Reflexion  
& $61.2$ &$ 15.7$ & $106$ & $0.51$ & $41.0$ &$ 8.0$ & $153$ & $0.46$ & $25.0$ & $7.4$ & $13$ & $0.66$ \\

&  Deliberation    & Debate     
& $17.9$ & $\mathbf{10.1}$ & $569$  & $0.03$ & $35.0$ &$ 8.8$ &$ 522 $& $0.12$ & $18.5$ &$ 6.7$ & $90$ & $0.07$ \\

& Deliberation & CER        
& $34.3$ & $15.2$ & $237 $ & $0.13 $& $38.0 $&$ 8.1$ & $255$ & $0.26$ & $4.0$ & $6.3 $& $44$ & $0.03$ \\

&Deliberation       & Traj-Boost 
& $59.7$ & $15.8 $& $279 $&$ 0.19$ & $43.0 $&$ 7.6$ & $80$ & $0.93$ & $11.0$ & $6.0 $& $10$ & $0.38$ \\

& Critic    & Retrospex  
& $22.4$ &$ 15.3$ &$ 121$ & $0.16 $& $11.0$ & $6.8$ & $67$ & $0.28$ & $13.0$ & $7.6$ & $9$ & $0.49$ \\

&Critic   & LAC        
& $44.0$ &$ 17.0$ &$ 381$ & $0.10$ & $\mathbf{60.0}$ & $10.8$ & $281$ & $0.37$ & $22.0$ & $6.8$ & $87$ &$ 0.09$ \\
& \cellcolor{ourslavender}\textbf{Critic}
& \cellcolor{ourslavender}\textbf{Ours (SAG)}
& \cellcolor{ourslavender}$\mathbf{63.0}$
& \cellcolor{ourslavender}$11.0$
& \cellcolor{ourslavender}$51$
& \cellcolor{ourslavender}$\mathbf{1.08}$
& \cellcolor{ourslavender}$\mathbf{60.0}$
& \cellcolor{ourslavender}$8.0$
& \cellcolor{ourslavender}$69$
& \cellcolor{ourslavender}$\mathbf{1.49}$
& \cellcolor{ourslavender}$\mathbf{36.5}$
& \cellcolor{ourslavender}$\mathbf{5.2}$
& \cellcolor{ourslavender}$\mathbf{6}$
& \cellcolor{ourslavender}$\mathbf{2.08}$ \\

\midrule
\multirow{8}{*}{\rotatebox{90}{\shortstack{\textbf{Mistral-}\\\textbf{7B-Instruct}}}}
& Basic       & ReAct      
& $30.6$ & $15.4$ & $\mathbf{56}$ &$ 1.00 $& $38.0 $& $7.8$ & $\mathbf{54}$& $1.00$ & $30.5$ & $7.3 $& $\mathbf{11}$ & $1.00$ \\
& Deliberation  & Reflexion  
& $46.3 $&$ 15.9$ &$ 145$ &$ 0.58$ &$ 36.0$ & $6.7$& $169 $& $0.30 $&$ 24.0$ &$ 7.4 $&$ 17 $& $0.51 $\\
& Deliberation      & Debate     
& $20.2$ & $12.9 $&$ 655$ &$ 0.06$ & $30.0 $& $9.1$ & $542$ & $0.08$ & $12.0$ & $6.9$ & $125$ &$ 0.03$ \\

& Deliberation   & CER        
& $43.3$ &$ 13.5$ & $247$ &$ 0.32$ &$ 47.0$ & $\mathbf{6.3}$ &$ 222 $& $0.30$ & $8.0 $& $6.2$ & $18$ & $0.16$ \\

& Deliberation          & Traj-Boost 
& $30.6$ & $16.2 $& $334$ & $0.17$ & $42.0$ & $7.3$ & $83$ &$ 0.72 $& $13.5$ &$ 5.7$ &$ 16 $& $0.31 $\\

& Critic    & Retrospex  
& $20.2 $&$ 11.9 $&$ 146$ & $0.25 $& $31.0$ & $8.1$ & $58$ & $0.75$ & $7.0$ & $9.3$ & $14$ & $0.28$ \\

&Critic    & LAC        
& $41.8$ & $17.5$ & $546$ & $0.14$ & $47.0$ &$ 11.1$ &$ 293$ & $0.23$ & $19.5$ & $7.1$ & $75$ &$ 0.09$ \\
& \cellcolor{ourslavender}\textbf{Critic }
& \cellcolor{ourslavender}\textbf{Ours (SAG)}
& \cellcolor{ourslavender}$\mathbf{53.2}$
& \cellcolor{ourslavender}$\mathbf{11.7}$
& \cellcolor{ourslavender}$61$
& \cellcolor{ourslavender}$\mathbf{1.61}$
& \cellcolor{ourslavender}$\mathbf{48.0}$
& \cellcolor{ourslavender}$7.9$
& \cellcolor{ourslavender}$58$
& \cellcolor{ourslavender}$\mathbf{1.18}$
& \cellcolor{ourslavender}$\mathbf{38.0}$
& \cellcolor{ourslavender}$\mathbf{5.6}$
& \cellcolor{ourslavender}$\mathbf{11}$
& \cellcolor{ourslavender}$\mathbf{1.26}$ \\
\bottomrule
\end{tabular}}
\vspace{-0.3cm}
\end{table}

\vspace{-0.3cm}

\begin{wrapfigure}{r}{0.42\textwidth}
    \centering
    \vspace{-0.5cm}
    \includegraphics[width=0.42\textwidth]{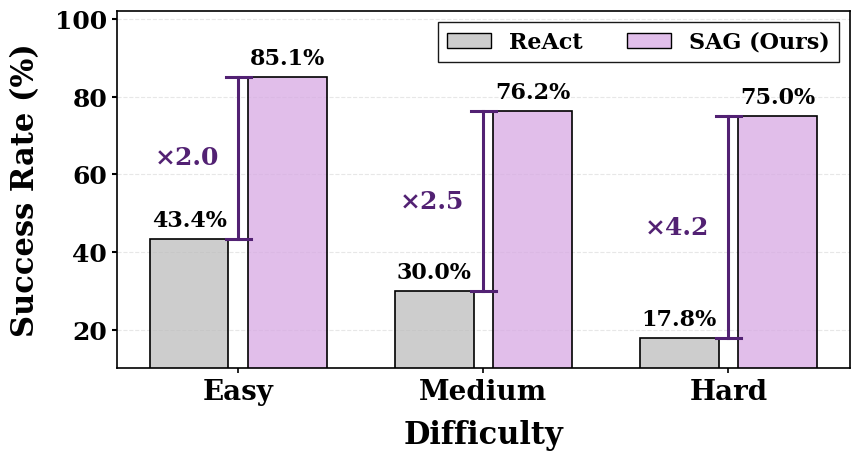}
    \caption{Success rate over task level. Task difficulty is stratified by the trajectory length of a GPT-4.1 ReAct agent.}
    \label{fig:level}
    \vspace{-0.5cm}
\end{wrapfigure}
\section{Experimental Results}
\vspace{-0.3cm}
\label{sec:main_results}
\subsection{Selective critique achieves strong success at near-ReAct cost}
\vspace{-0.3cm}

Table~\ref{tab:main_results} summarizes the performance across methods. Overall, deliberation and always-on critique tend to improve success at the cost of substantially increased token usage, whereas SAG achieves a more favorable balance by selectively invoking critique only when needed.

\textbf{SAG improves long-horizon success while preserving a ReAct-like token budget} Across all backbone--benchmark settings, SAG improves success over ReAct in every case, with an average relative improvement of $91\%$ while keeping token usage close to ReAct on ALFWorld/WebShop and far below always-on critique on BabyAI-Text. This yields higher normalized token efficiency, especially on longer-horizon environments such as ALFWorld. Notably, in several settings, SAG with a 7B backbone attains success rates close to those of a stronger GPT-4.1-based ReAct agent, narrowing the gap through improved decision quality rather than increased model capacity.

\begin{figure*}[t]
\vspace{-1.5cm}    
\centering
    \includegraphics[width=1\linewidth]{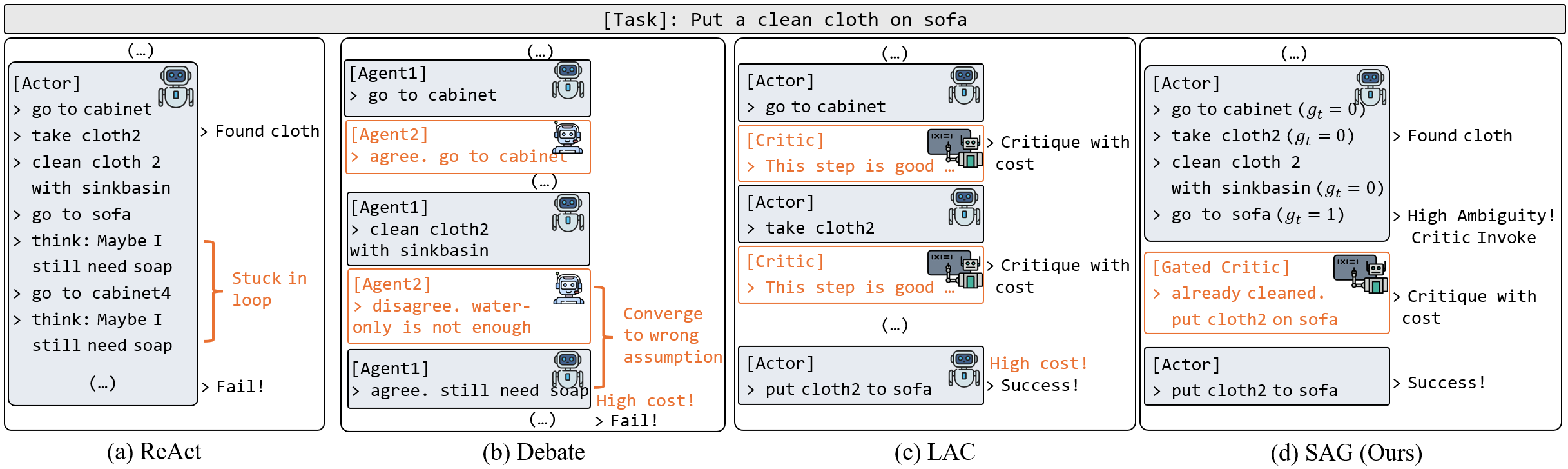}
    \vspace{-0.5cm}
    \caption{Case Study: SAG effectively corrects the trajectory, avoiding both failure and pervasive critique overhead.}
    \label{fig:casestudy}
    \vspace{-0.4cm}
\end{figure*}

\vspace{-0.3cm}
\paragraph{Always-on critique and deliberation tend to be token-inefficient}
Methods that invoke critique or deliberation at most steps quickly become cost-dominated, increasing token usage by $3\times$ to over $10\times$ relative to ReAct while yielding only modest success gains. As shown in Figure~\ref{fig:alfworld_tradeoff_qwen}, they shift rightward on the performance--cost plane, reflecting higher cost without proportional reliability gains. Qualitative analysis (Figure~\ref{fig:casestudy}) shows that always-on critique repeatedly intervenes at low-risk states, accumulating redundant confirmations and growing context length, whereas SAG corrects a few high-risk decisions while avoiding unnecessary feedback.

\begin{wrapfigure}{r}{0.48\textwidth}
    \centering
    \vspace{-0.5cm}
    \includegraphics[width=0.48\textwidth]{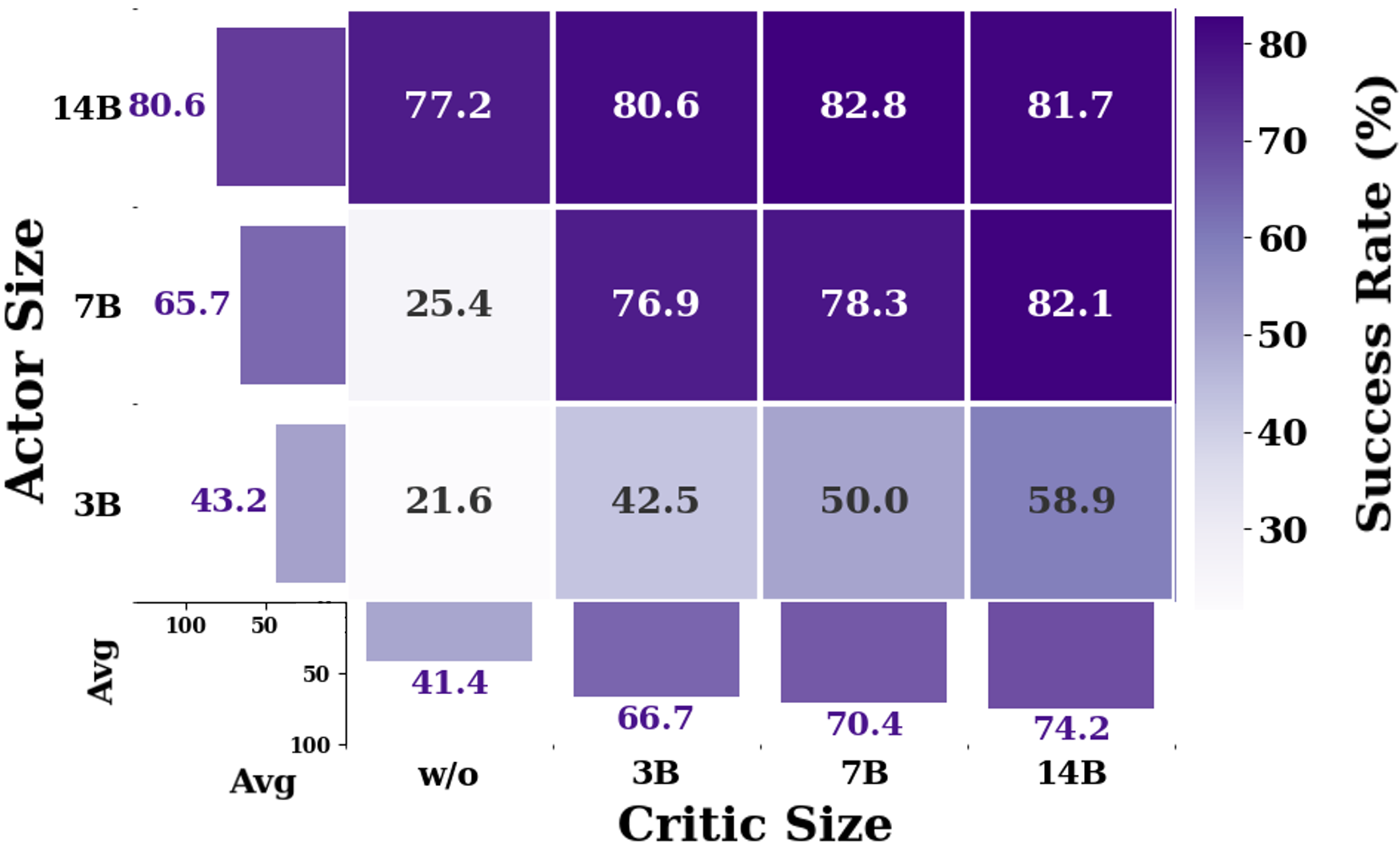}
    \caption{Actor--critic capacity scaling for SAG. Each cell reports success rate for a given actor--critic size pairing. Marginal bars show mean performance across actor and critic sizes.}
    \label{fig:scale_heatmap}
    \vspace{-0.5cm}
\end{wrapfigure}
\textbf{SAG exhibits increasing advantages on harder episodes}  
Figure~\ref{fig:level} analyzes performance as a function of episode difficulty. We stratify test episodes into easy, medium, and hard subsets based on the trajectory length of a GPT-4.1 ReAct agent, with longer trajectories indicating higher difficulty. We compare SAG and ReAct on the same subsets. SAG consistently outperforms ReAct across all difficulty levels, with the performance gap widening as difficulty increases. On easy episodes, SAG improves success from $43.4\%$ to $85.1\%$ (approximately $2\times$ relative improvement), whereas on hard episodes it improves success from $17.8\%$ to $75.0\%$, corresponding to over a $4\times$ improvement. This trend indicates that selective critique is particularly effective in challenging long-horizon scenarios, where early mistakes are more likely to propagate and dominate outcomes.

\vspace{-0.2cm}
\subsection{Why Gated Critique Works: Capacity Scaling and Ablations}
\vspace{-0.2cm}

\begin{wrapfigure}{r}{0.48\textwidth}
\vspace{-0.4cm}
    \begin{minipage}{0.48\textwidth}
        \centering
        \scriptsize
        \setlength{\tabcolsep}{4pt}
        \renewcommand{\arraystretch}{1.15}
        \captionof{table}{Ablation study on ALFWorld. Global indicates using global ambiguity $\bar H_t$; Local indicates using local ambiguity $m_t$. Always removes the gate and invokes critique at every step; w/o critic disables critic invocation.}
        \label{tab:ablation_alfworld}
        \resizebox{\linewidth}{!}{%
        \begin{tabular}{cc lccc}
        \toprule
        \textbf{Global} & \textbf{Local} & \textbf{Variant}
        & \textbf{Succ} $\uparrow$ & \textbf{\#Horizon} $\downarrow$ & \textbf{\#Token} $\downarrow$ \\
        \midrule
        \rowcolor{oursgray}
        \ding{51} & \ding{51} & \textbf{Full (SAG)} & \textbf{78.3} & 11.4 & 56 \\
        \ding{55} & \ding{51} & w/o $\bar H_t$      & 77.6 & 12.1 & 51 \\
        \ding{51} & \ding{55} & w/o $m_t$           & 74.6 & 11.4 & 52 \\
        \ding{55} & \ding{55} & Always (no gating)  & 62.0 & 17.4 & 282 \\
        \ding{55} & \ding{55} & w/o critic          & 33.0 & 19.3 & 50 \\
        \bottomrule
        \end{tabular}%
        }
        \vspace{-0.3cm}
    \end{minipage}
\end{wrapfigure}
\paragraph{Effect of actor and critic capacity}
Figure~\ref{fig:scale_heatmap} analyzes SAG across actor--critic capacity pairings (3/7/14B) and a no-critic setting. Actor scaling provides the largest gains: without critique, success increases from $21.6\% \rightarrow 25.4\% \rightarrow 77.2\%$ from 3B to 14B, indicating that actor capacity is the primary driver of long-horizon performance. Nevertheless, gated critique consistently improves all actor sizes. Notably, a 7B actor with a 3B critic already matches a 14B actor without critique ($76.9\%$ vs.\ $77.2\%$), showing that modest critic capacity can close much of the gap to a larger actor. These gains are strongest for smaller actors and saturate as the actor improves, suggesting that selective critique complements actor capacity rather than replacing it. Extended 32B and always-on comparisons are in Appendix~\ref{app:extended_scaling}.

\vspace{-0.2cm}
\paragraph{Ablation study}
Table~\ref{tab:ablation_alfworld} ablates the two ambiguity signals—global entropy $\bar H_t$ and local ambiguity $m_t$—as well as the gating mechanism in SAG. Removing either signal leads to a small but consistent drop in success, from $78.3\%$ to $77.6\%$ without $\bar H_t$ and to $74.6\%$ without $m_t$, indicating that global and local ambiguity provide complementary cues for identifying high-risk states. In contrast, removing the gate and invoking critique at every step reduces success to $62.0\%$ while increasing token usage to $282$k tokens, showing that indiscriminate critique can be both costly and harmful. Finally, disabling critic invocation lowers success to $33.0\%$, confirming that SAG's gains come from selectively deploying critique at ambiguous steps rather than from self-improvement alone.

\begin{wrapfigure}{r}{0.45\textwidth}
\vspace{-0.45cm}
    \begin{minipage}{0.45\textwidth}
        \centering
        \scriptsize
        \setlength{\tabcolsep}{4pt}
        \renewcommand{\arraystretch}{1.15}
        \captionof{table}{Matched online-SFT controls. All methods use the same online update protocol.}
        \label{tab:matched_sft_alfworld}
        \vspace{-0.1cm}
        \resizebox{\linewidth}{!}{%
        \begin{tabular}{lccc}
        \toprule
        \textbf{Method} & \textbf{Critique} & \textbf{Succ.} $\uparrow$ & \textbf{\#Token} $\downarrow$ \\
        \midrule
        ReAct + online SFT & None & 30.2 & 57 \\
        LAC + online SFT & Always-on & 74.2 & 341 \\
        \rowcolor{oursgray}
        \textbf{SAG} & Gated & \textbf{78.3} & \textbf{56} \\
        \bottomrule
        \end{tabular}%
        }
        \vspace{-0.3cm}
    \end{minipage}
\end{wrapfigure}

\vspace{-0.3cm}
\paragraph{Selective critique is not explained by self-improvement alone}
Table~\ref{tab:matched_sft_alfworld} compares SAG with matched online-SFT controls. ReAct + online SFT uses the same update protocol without critique, but improves success only to $30.2\%$. LAC + online SFT uses always-on critique and reaches $74.2\%$, but requires $341$k tokens. SAG achieves higher success ($78.3\%$) with only $56$k tokens, showing that the gain comes from selectively deploying critique rather than online SFT alone. Full results on all benchmarks are in Appendix~\ref{app:self_improve_controls}.

\begin{wrapfigure}{r}{0.52\textwidth}
\vspace{-0.6cm}
    \begin{minipage}{0.52\textwidth}
        \centering
        \includegraphics[width=\linewidth]{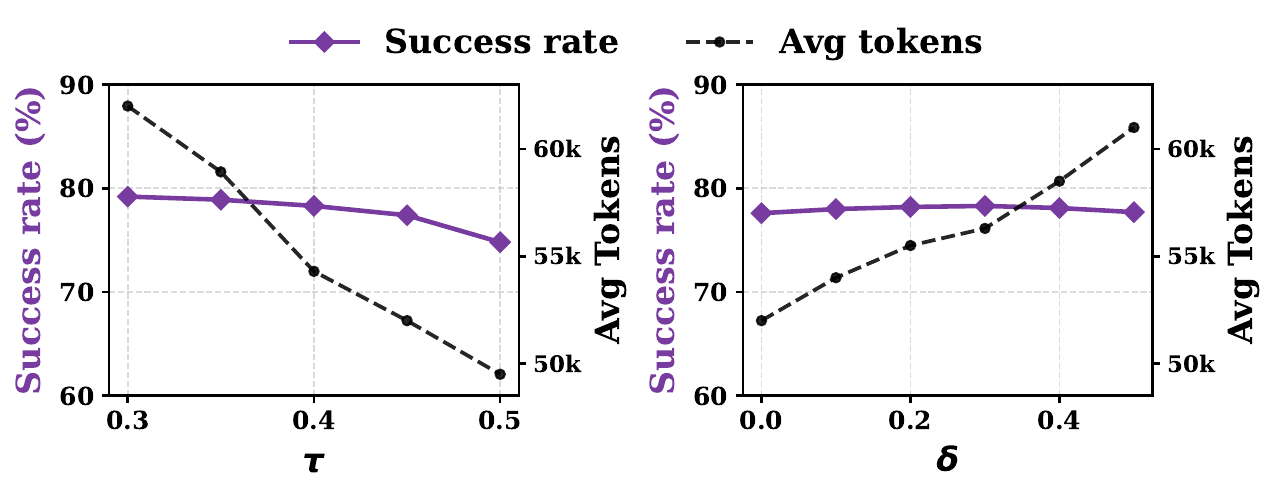}
        \vspace{-0.7cm}
        \caption{Parameter sensitivity analysis}
        \label{fig:sensitivity}
        \vspace{-0.4cm}
    \end{minipage}
\end{wrapfigure}
\vspace{-0.3cm}
\paragraph{Parameter sensitivity}
Figure~\ref{fig:sensitivity} shows that SAG is robust to the gate thresholds. Increasing $\tau$ smoothly trades success for lower token cost: over $\tau\in[0.30,0.45]$, success changes mildly while token cost drops from $62$k to $52$k. Varying $\delta$ has limited impact on success ($77.6\%$--$78.3\%$), though larger values invoke critique more often. Overall, SAG does not require fine-grained threshold tuning. Importantly, the same threshold values are used across all benchmarks and backbone models in Table~\ref{tab:main_results} without task-specific tuning. Detailed threshold sweeps are reported in Appendix~\ref{app:threshold_sensitivity}.

\vspace{-0.3cm}
\subsection{Learning Dynamics of Self-Improving Agents}
\begin{wrapfigure}{r}{0.53\textwidth}
\vspace{-0.6cm}
    \begin{minipage}{0.53\textwidth}
        \centering
        \begin{subfigure}[t]{0.49\linewidth}
            \centering
            \includegraphics[width=\linewidth]{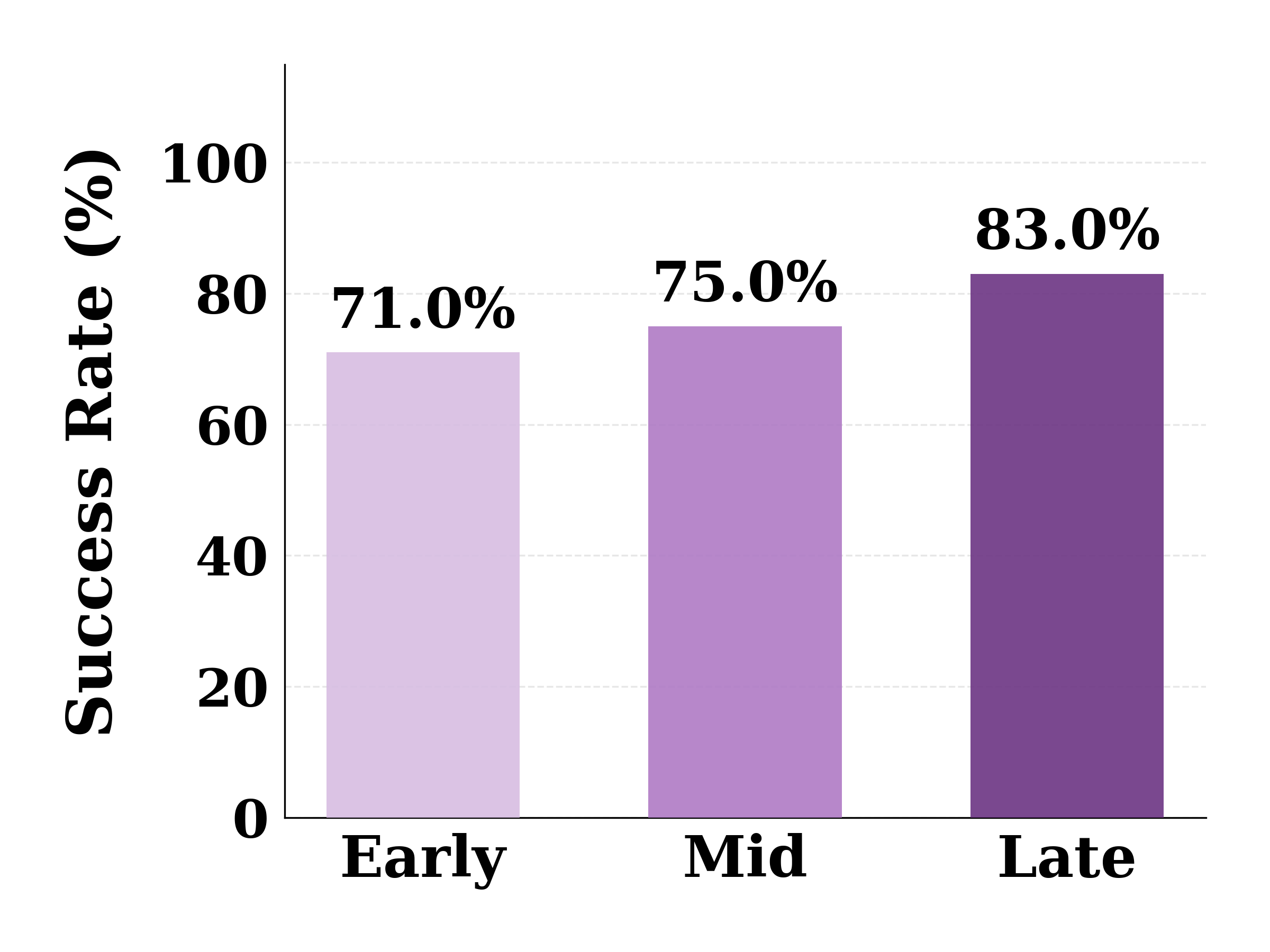}
            \vspace{-0.6cm}
            \caption{Success rate (\%)}
            \label{fig:learn_succ}
        \end{subfigure}
        \hfill
        \begin{subfigure}[t]{0.49\linewidth}
            \centering
            \includegraphics[width=\linewidth]{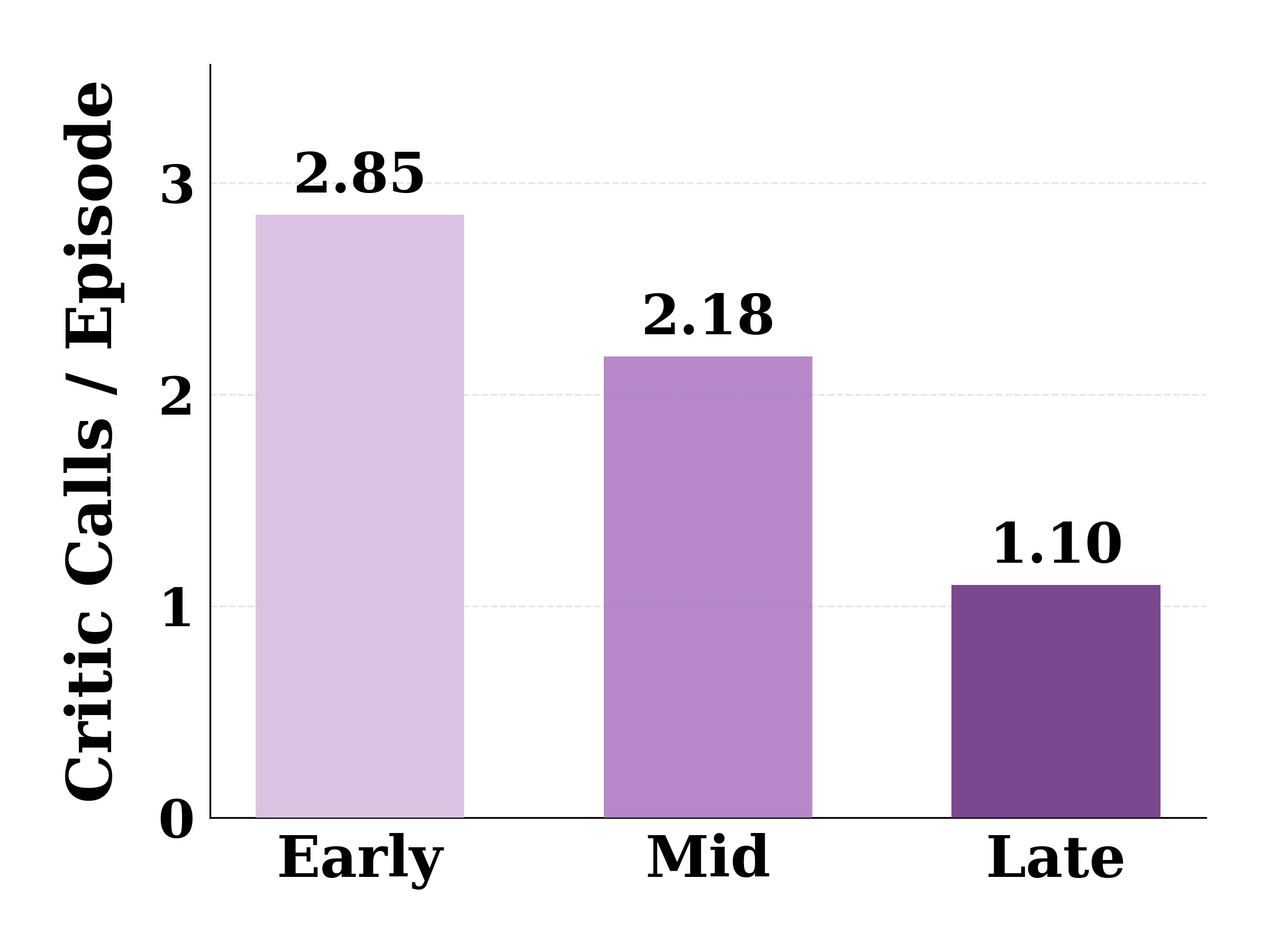}
            \vspace{-0.6cm}
            \caption{Critic calls / episode}
            \label{fig:learn_critic}
        \end{subfigure}

        \begin{subfigure}[t]{0.49\linewidth}
            \centering
            \includegraphics[width=\linewidth]{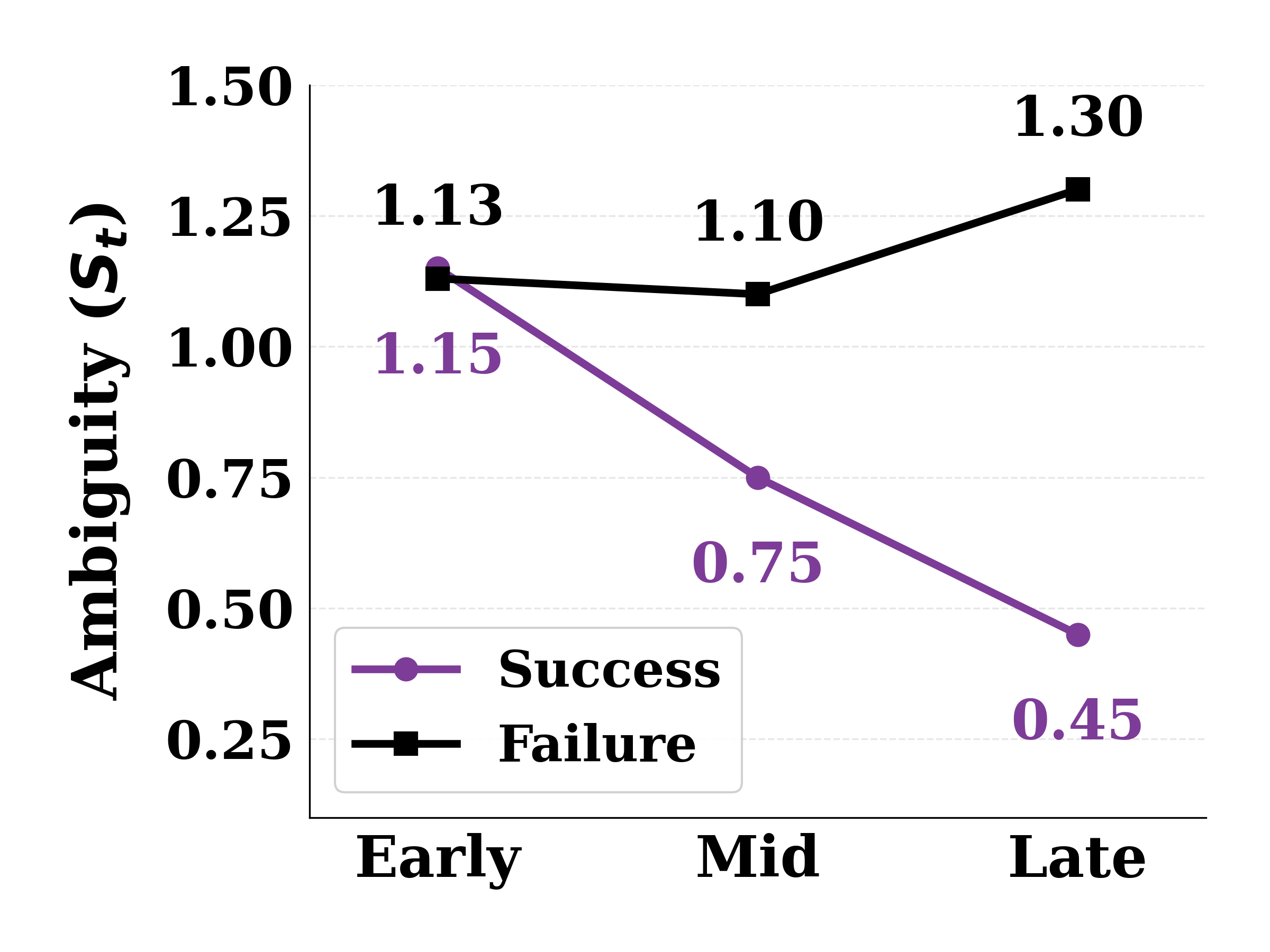}
            \vspace{-0.6cm}
            \caption{$S_t$ for success vs.\ failure}
            \label{fig:learn_succfail}
        \end{subfigure}
        \hfill
        \begin{subfigure}[t]{0.49\linewidth}
            \centering
            \includegraphics[width=\linewidth]{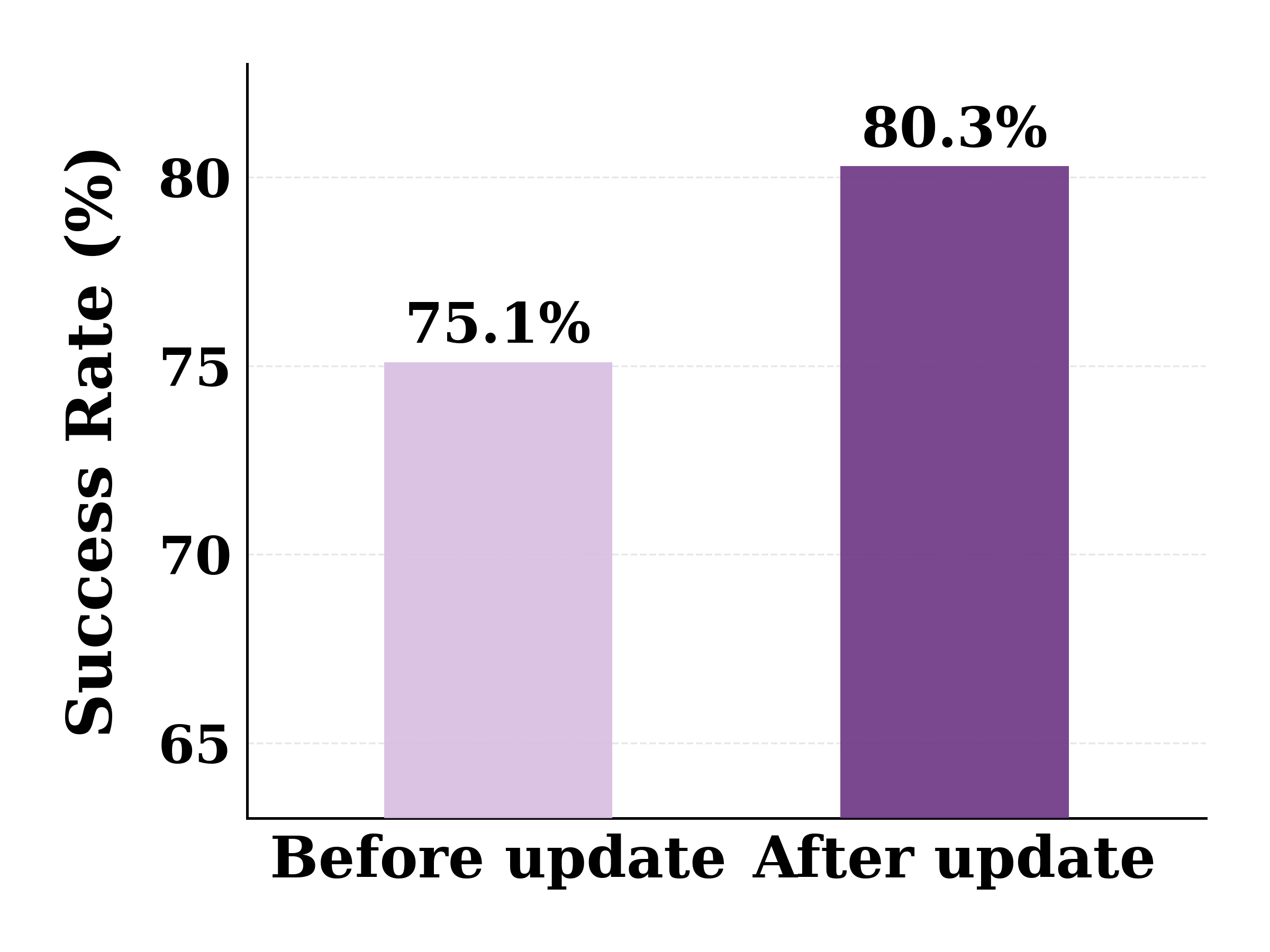}
            \vspace{-0.6cm}
            \caption{Seen: before vs.\ after}
            \label{fig:learn_beforeafter}
        \end{subfigure}

        \vspace{-0.2cm}
        \caption{Learning dynamics of SAG under online self-improvement. Metrics are aggregated into Early/Mid/Late phases over 134 evaluation episodes.}
        \label{fig:learning_dynamics}
        \vspace{-0.5cm}
    \end{minipage}
\end{wrapfigure}
\vspace{-0.2cm}
\paragraph{Online self-improvement reduces reliance on critique}
Figure~\ref{fig:learn_succ} and Figure~\ref{fig:learn_critic} show that, as interaction proceeds, SAG's success rate increases ($71.0\% \rightarrow 75.0\% \rightarrow 83.0\%$) while critic calls decrease ($2.85 \rightarrow 2.18 \rightarrow 1.10$). This suggests that successful critic-assisted trajectories are distilled into the actor, making it more confident under the same gate and less reliant on critique.

\vspace{-0.2cm}
\paragraph{Learning reduces decision risk rather than increasing computation}
Figure~\ref{fig:learn_succfail} shows that ambiguity remains well separated between successes and failures as learning progresses.
For successful episodes, ambiguity consistently decreases ($1.15 \rightarrow 0.75 \rightarrow 0.45$), whereas failures exhibit persistently higher and increasing ambiguity ($1.13 \rightarrow 1.10 \rightarrow 1.30$), indicating that remaining errors concentrate on harder decision points.
As a diagnostic analysis excluded from the main success metric, re-evaluating previously encountered episodes after online updates shows improved performance (Figure~\ref{fig:learn_beforeafter}), increasing from $75.1\%$ to $80.3\%$.
Together, these results show that critique acts as a transient scaffold: as learning progresses, SAG internalizes effective behaviors and shifts decision-making toward lower-risk states without increasing inference-time computation.

\vspace{-0.4cm}
\section{Conclusion}
\label{sec:discussion}
\vspace{-0.3cm}
We introduced SAG, a cost-aware framework that selectively invokes critique to improve the reliability of LLM agents in long-horizon decision-making. SAG formulates critique invocation as a step-wise decision problem and employs a lightweight, training-free gate based on action-level ambiguity. Across multiple long-horizon benchmarks and backbone models, SAG consistently achieves a superior performance--cost trade-off compared to both no-critique and always-on critique agents. These results suggest that selective critique provides a practical and scalable pathway for improving the reliability and efficiency of LLM agents in complex sequential environments. 

\section*{Acknowledgments}
\vspace{-0.3cm}
{\small
This work was partly supported by the Institute of Information \& Communications Technology Planning \& Evaluation (IITP) grant (No. RS-2026-25519475), the National Research Foundation of Korea (NRF) grants (RS-2023-00278812, RS-2025-02214082, RS-2026-25522801) funded by the Korean government (MSIT), and Korea Institute of Police Technology (KIPoT) funded by the Korean National Police Agency \& Korea Customs Service (RS-2026-25536784). The computing resources were partly supported by Institute of Information \& communications Technology Planning \& Evaluation (IITP, AI Computing Support Project for R\&D) grant funded by the Korea government(MSIT) (RS-2026-25505492)}

\bibliography{ref,BMIL_260925}
\bibliographystyle{unsrt}

\newpage

\par\noindent\rule{\textwidth}{2pt}
\begin{table}[h]
    \vspace{-0.3cm}
    \centering
    \Large
    \begin{tabular}{p{0.95\textwidth}}
    \centering
    \textbf{Appendix: Selective Critique for Cost-Aware LLM Agents in Long-Horizon Decision Making}\\
    \end{tabular}
    \vspace{-.5cm}
\end{table}
\par\noindent\rule{\textwidth}{1pt}
\appendix
\vspace{-0.4cm}
\tableofcontents   
\addtocontents{toc}{\protect\setcounter{tocdepth}{2}}
\setcounter{section}{0}

\newcommand{\appline}[3]{%
  \noindent\textbf{#1}\hspace{0.6em}#2\dotfill\pageref{#3}\par
}
\newcommand{\appsubline}[3]{%
  \noindent\hspace*{1.7em}\textbf{#1}\hspace{0.6em}#2\dotfill\pageref{#3}\par
}
\newcommand{\appsubsubline}[3]{%
  \noindent\hspace*{3.2em}\textbf{#1}\hspace{0.6em}#2\dotfill\pageref{#3}\par
}

\clearpage
\normalsize

\section{Broader Impacts and Limitations}
\label{Impact}
\paragraph{Broader Impacts}
This work contributes a practical methodology for improving the reliability and efficiency of LLM-based agents, a capability that is increasingly important as such systems are deployed in complex interactive settings. Because SAG is lightweight, training-free, and compatible with existing agent frameworks, it can be readily integrated into current LLM-agent pipelines used in both academic research and industrial applications. In research contexts, SAG provides a simple yet principled mechanism for studying cost-aware reasoning allocation and selective deliberation in long-horizon decision-making. In applied settings, the framework can reduce unnecessary token usage and latency while preserving reliability, making agent-based systems more economically and computationally sustainable. As with other general-purpose LLM agent techniques, the societal impact ultimately depends on how such systems are deployed; responsible development and application-level safeguards remain important for ensuring beneficial outcomes.

\paragraph{Limitations}
While SAG demonstrates consistent improvements across multiple environments and backbone models, several aspects remain open for further exploration. First, our current gating mechanism relies on ambiguity signals derived from the actor’s action distribution; alternative uncertainty or value-estimation signals may further improve critique allocation. Second, the current study focuses on representative long-horizon interactive benchmarks, and evaluating SAG in additional real-world agent settings would provide further insight into its generality. Finally, the present work uses a fixed critic architecture; future work may explore adaptive or learned critics that further enhance the reliability–efficiency trade-off.



\section{Notation Table}
\label{app:notation}

\begin{table}[h]
\centering
\small
\begin{tabular}{ll}
\toprule
\textbf{Notation} & \textbf{Description} \\
\midrule
$t$ & Time step in an interaction episode \\

$o_t$ & Environment observation at step $t$ \\

$x_t$ & Actor context at step $t$, including task instruction, observation, and truncated history \\

$\mathcal{A}_t$ & Set of admissible actions at step $t$ \\

$a_t$ & Action executed at step $t$ \\

$r_t$ & Actor reasoning state generated from $x_t$ at step $t$ \\

$\pi_\theta$ & Actor language-model policy parameterized by $\theta$ \\

$K$ & Actor policy update interval \\

$\ell_t(a)$ & Length-normalized token log-likelihood score of admissible action $a$ under the actor \\

$q_t(a)$ & Softmax-normalized actor-induced distribution over admissible actions \\

$a_1, a_2$ & Top-1 and top-2 actions under $q_t$ \\

$q_{\max}$ & Maximum probability mass under $q_t$, i.e., $q_t(a_1)$ \\

$g_t \in \{0,1\}$ & Binary critic-invocation decision at step $t$ \\

$c$ & Computational cost of invoking the critic \\

$\pi_\phi$ & Critic policy parameterized by $\phi$ \\

$\bar H_t$ & Normalized entropy of $q_t$ (global ambiguity)\\

$m_t$ & Log-probability margin between top-1 and top-2 actions (local ambiguity) \\

$S_t$ & Gating score combining global and local ambiguity\\

$\tau$ & Gating threshold controlling overall critic usage \\

$\delta$ & Margin threshold for detecting near-boundary decisions \\

$\Delta_t$ & Step-level improvement attributable to critique \\

$z_t=(x_t,r_t)$ & Decision state after actor reasoning at step $t$ \\

$\mathrm{VoI}(z_t)$ & Value of information of invoking critique at decision state $z_t$ \\

$\mathrm{err}_t$ & Event that the actor's immediate decision at step $t$ is incorrect \\

$\rho$ & Probability that the critic corrects an incorrect actor decision \\

$\gamma$ & Average gain from a successful critic correction \\

$\mathcal{D}$ & Buffer of successful context--action pairs for online self-improvement \\

$\mathcal{L}_{\mathrm{SFT}}$ & Supervised fine-tuning loss used for online self-improvement \\

\bottomrule
\end{tabular}
\label{tab:notation}
\end{table}

\newpage
\section{Proof of Lemma~\ref{lem:voi_gate}}
\label{app:voi_gate}

We consider the one-step decision problem at decision state $z_t=(x_t,r_t)$, where the agent chooses whether to invoke the critic. Let $g_t \in \{0,1\}$ denote the invocation decision. The Bayes-optimal decision maximizes the conditional expected net utility given $z_t$. Invoking the critic ($g_t=1$) yields a random improvement $\Delta_t$ (e.g., increase in success probability or expected return) and incurs a deterministic cost $c>0$. If the critic is not invoked ($g_t=0$), the improvement is defined as $0$ and no additional cost is incurred.

Conditioned on $z_t$, the expected net utility of choosing $g_t=1$ is
\begin{equation}
\mathbb{E}[\Delta_t \mid z_t] - c,
\end{equation}
whereas the expected net utility of choosing $g_t=0$ is $0$. Therefore, invoking the critic is optimal iff the expected net gain is positive:
\begin{equation}
g_t^\star=\mathbb{I}\left[\mathbb{E}[\Delta_t \mid z_t] - c > 0\right]=\mathbb{I}\left[\mathbb{E}[\Delta_t \mid z_t] > c\right].
\end{equation}
Recalling the definition $\mathrm{VoI}(z_t) := \mathbb{E}[\Delta_t \mid z_t]$ completes the proof. \hfill$\square$

\section{Proof of Proposition~\ref{prop:entropy_err}}
\label{app:entropy_bounds}

Fix a step $t$ and decision state $z_t=(x_t,r_t)$, and let $q_t$ be the actor-induced distribution over admissible actions $\mathcal{A}_t$. Let $a_1=\arg\max_{a\in\mathcal{A}_t} q_t(a)$ and $q_{\max}=q_t(a_1)$.

\paragraph{Step 1: Error probability under the belief interpretation.}
Equivalently, Eq.~\eqref{eq:err_eq_1minusqmax} holds if we define $\Pr(\text{$a$ is correct}\mid z_t):=q_t(a)$ over $\mathcal{A}_t$.
Under this interpretation, the probability that the greedy choice $a_1$ is incorrect equals the probability mass assigned to all other actions:
\begin{equation}
\Pr(\mathrm{err}_t \mid z_t)=1-\Pr(\text{$a_1$ is correct}\mid z_t)=1-q_{\max},
\end{equation}
which proves Eq.~\eqref{eq:err_eq_1minusqmax}.

\paragraph{Step 2: $-\log q_{\max}$ lower bounds entropy.}
Let $H(q_t)=-\sum_{a\in\mathcal{A}_t} q_t(a)\log q_t(a)$ denote the entropy. Since $q_{\max} \ge q_t(a)$ for all $a\in\mathcal{A}_t$, we have $\log q_{\max} \ge \log q_t(a)$ and thus $-\log q_t(a) \ge -\log q_{\max}$ for all $a$. Multiplying by $q_t(a)\ge 0$ and summing over $a$ yields
\begin{equation}
H(q_t) = \sum_{a\in\mathcal{A}_t} q_t(a)\big(-\log q_t(a)\big) \ge \sum_{a\in\mathcal{A}_t} q_t(a)\big(-\log q_{\max}\big)=-\log q_{\max}.
\end{equation}
Exponentiating both sides gives $q_{\max} \ge e^{-H(q_t)}$.

Combining with Step 1, we obtain
\begin{equation}
\Pr(\mathrm{err}_t \mid z_t) = 1-q_{\max} \le 1-e^{-H(q_t)}.
\end{equation}
Finally, using the normalized entropy definition $\bar H_t = H(q_t)/\log|\mathcal{A}_t|$ (for $|\mathcal{A}_t|\ge 2$), we have
$e^{-H(q_t)} = e^{-\bar H_t \log|\mathcal{A}_t|} = |\mathcal{A}_t|^{-\bar H_t}$, which yields
\begin{equation}
\Pr(\mathrm{err}_t \mid z_t) \le 1-|\mathcal{A}_t|^{-\bar H_t}.
\end{equation}
This completes the proof.
\hfill$\square$

\newpage
\section{Proof of Proposition~\ref{prop:margin_top2}}
\label{app:margin_risk}

Fix a step $t$ and decision state $z_t=(x_t,r_t)$. Let $q_t$ be the actor-induced distribution over admissible actions $\mathcal{A}_t$. Let $a_1$ and $a_2$ denote the top-1 and top-2 actions under $q_t$, and write $q_1 := q_t(a_1)$ and $q_2 := q_t(a_2)$. Recall the log-probability margin $m_t := \log q_1 - \log q_2 = \log(q_1/q_2)$.

Assume the correct action lies in $\{a_1,a_2\}$. Conditioned on this event (denoted $\mathrm{top\text{-}2}$), the induced conditional distribution over $\{a_1,a_2\}$ is
\begin{equation}
\Pr(\text{$a_i$ is correct} \mid z_t,\mathrm{top\text{-}2}) = \frac{q_i}{q_1+q_2},\quad i\in\{1,2\}.
\end{equation}
Therefore,
\begin{equation}
\Pr(\mathrm{err}\mid z_t,\mathrm{top\text{-}2})=\Pr(\text{$a_2$ is correct} \mid z_t,\mathrm{top\text{-}2})=\frac{q_2}{q_1+q_2}.
\end{equation}
Rewriting in terms of the margin gives
\begin{equation}
\Pr(\mathrm{err}\mid z_t,\mathrm{top\text{-}2})=\frac{q_2}{q_1+q_2}=\frac{1}{1+q_1/q_2}=\frac{1}{1+e^{m_t}}.
\end{equation}
Finally, since $e^{m_t}>0$, we use $1+e^{m_t}\ge e^{m_t}$ to obtain
\begin{equation}
\frac{1}{1+e^{m_t}}\le\frac{1}{e^{m_t}}=e^{-m_t}.
\end{equation}
This proves Eq.~\eqref{eq:margin_err}.
\hfill$\square$

\newpage
\section{Prompts}
\label{app:prompt}
\begin{table*}[hbt!]
\tiny
\centering
\caption{Prompts used for ALFWorld}
\footnotesize
\begin{tabular}{@{}l@{}}
\toprule
\begin{minipage}{0.95\linewidth}
\raggedright
\fontfamily{qcr}\selectfont
\textbf{[Actor Prompt]} 
You are an agent for household tasks. \\
Your goal is to complete the task described in natural language by reasoning step-by-step and producing an action. \\
Always output in the format: \\
"think: reasoning here" \\
"action: [chosen action]" \\
Actions must come from the admissible set provided. Do not invent new actions. \\

\textbf{[Critic Prompt]} 
You are a critic for an LLM-based agent. \\
Your GOAL: detect and correct hallucinations, loops, or inconsistent actions. \\
INPUTS: \\
- task: the goal \\
- history: what happened so far \\
- actions: valid moves. \\ 
INSTRUCTIONS: \\
1. verify objects: does the object the agent is targeting MATCH the task description? \\
2. verify history: did the agent already check this empty location or repeat failures? \\
3. decision: \\
   - If the agent is safe and correct, output "next\_action": "continue". \\
   - If the agent is wrong, provide a specific "next\_action" to fix it. \\
4. reasoning: \\
   - Also provide a short explanation in "reasoning". \\
   - The reasoning should explain why the proposed action is appropriate. \\
   - Keep it concise, one or two sentences. \\
\end{minipage}\\
\bottomrule
\end{tabular}
\end{table*}

\begin{table*}[hbt!]
\tiny
\centering
\caption{Prompts used for BabyAI-Text}
\footnotesize
\begin{tabular}{@{}l@{}}
\toprule
\begin{minipage}{0.95\linewidth}
\raggedright
\fontfamily{qcr}\selectfont
\textbf{[Actor Prompt]} 
You are an agent in the Grid world. \\
Your goal is to follow the natural language instruction step-by-step. \\
At each step, reason explicitly before acting. Use the format: \\
"think: reasoning here" \\
"action: [chosen action]" \\
Primitive actions include: go left, go right, go forward, pick up object, drop object, etc. \\
\textbf{[Critic Prompt]} 
You are a critic helping an actor agent navigate a small grid world. \\
Your GOAL: \\
- Detect when the actor is stuck, uncertain, or moving inefficiently.\\
- Propose a corrective plan that helps the actor make progress toward the goal.\\
INPUTS:\\
- goal: The current task the actor must complete.\\
- observation: The actor's current state in the grid world.\\
- valid actions: The list of actions the actor can execute now.\\
INSTRUCTIONS:\\
1. check progress:\\
- Determine whether the actor is making progress toward the goal.\\
- If the actor seems stuck in a loop, suggest actions that break the loop.\\
2. check validity:\\
- Only use actions from the valid actions list.\\
- Do not invent new actions.\\
3. proposed plan:\\
- Propose a SHORT sequence of 1 to 3 actions.\\
- The plan should be directly executable in the current state.\\
4. reasoning:\\
- Provide a brief reason explaining why the plan is helpful.\\
- Keep the reasoning concise, at most one sentence.\\
\end{minipage}\\
\bottomrule
\end{tabular}
\end{table*}

\begin{table*}[hbt!]
\tiny
\centering
\caption{Prompts used for WebShop}
\footnotesize
\begin{tabular}{@{}l@{}}
\toprule
\begin{minipage}{0.95\linewidth}
\raggedright
\fontfamily{qcr}\selectfont
\textbf{[Actor Prompt]} You are a shopping assistant. \\
Your task is to satisfy the given shopping goal. \\
At each step, reason aloud and then select one action. \\
Use \texttt{search[query]} for the initial query; after search results are shown, choose from the web actions available in the current observation, such as \texttt{click[item]}, \texttt{click[option]}, navigation, or \texttt{click[Buy Now]}. \\
Use format: \\
"think: reasoning here" \\
"action: [web action]" \\

\textbf{[Critic Prompt]} 
You are a critic for a WebShop shopping agent.\\
GOAL:\\
- Help the actor choose the best next action for completing the shopping task.\\
- Detect inefficient actions, loops, or choices that do not satisfy the user instruction.\\

INPUTS:\\
- user instruction: the shopping request, including constraints such as product type, size, flavor, count, price, rating, or other preferences.\\
- recent history: The recent actions and observations from the shopping trajectory.\\
- candidate actions: The list of actions currently available to the actor.\\

INSTRUCTIONS:\\
1. check constraints:\\
   - Identify the important constraints in the user instruction.\\
   - Prefer actions that move toward satisfying all constraints.\\

2. check progress:\\
   - Avoid repeating unhelpful actions.\\
   - Avoid going back unless it clearly helps recover from a mistake.\\
   - Prefer clicking the most relevant product, option, or navigation action.\\

3. choose action:\\
   - Choose exactly ONE best next action.\\
   - The action must be one of the candidate actions.\\
   - If no candidate action is clearly useful, choose the safest progress-making action.\\
   - Use "none" only when no valid progress-making action exists.\\

4. rationale:\\
   - Provide a very short reason explaining why the chosen action is appropriate.\\
   - Keep the rationale concise, one sentence at most.\\
\end{minipage}\\
\bottomrule
\end{tabular}
\end{table*}

\newpage
\section{Simulation Settings}
\label{app:setting}

\subsection{Benchmarks for Interactive Environments}
\label{app:benchmarks}

This section provides detailed descriptions of the interactive environments used in our experiments, including task configurations, scenario distributions, and evaluation protocols. These details are omitted from the main paper for brevity.

\textbf{ALFWorld}~\cite{shridhar2020alfworld} is a text-based household simulation environment built on top of ALFRED, where agents interact with a partially observed home environment through natural language actions. Each task consists of a high-level instruction (e.g., placing, cleaning, or heating objects) that must be decomposed into a long sequence of grounded actions such as navigation, object manipulation, and state changes. We follow the standard evaluation protocol, using the unseen test split to assess generalization to novel layouts and object configurations. Episodes typically require dozens of sequential actions, making performance sensitive to early mistakes and error accumulation.

\textbf{BabyAI-Text}~\cite{babyai} provides a natural-language interface to the BabyAI grid world, where agents must follow compositional instructions involving spatial relations, object attributes, and temporal ordering. Although the environment dynamics are deterministic and the observation space is compact, tasks are generated procedurally to ensure a wide range of unseen instruction combinations at test time. We evaluate on held-out instruction distributions to measure compositional generalization and multi-step planning accuracy independent of perceptual noise.

\textbf{WebShop}~\cite{yao2022webshop} is a web-based e-commerce environment that simulates realistic online shopping scenarios. Agents interact with dynamically generated product pages using text-based actions such as search queries, page navigation, and item selection. Each episode requires identifying and purchasing an item that satisfies multiple constraints specified in a natural-language goal. The environment presents noisy, unstructured textual information and numerous distractors, requiring agents to balance exploration cost and task success. For SAG, after the initial search step, $\mathcal{A}_t$ is constructed from the finite set of clickable actions exposed in the current page observation, such as visible products, item options, navigation controls, and \texttt{Buy Now}. Thus, the ambiguity gate and critic operate only over this finite candidate set rather than the open-ended space of possible search queries.

Across all environments, tasks require sustained reasoning and adaptive decision making over extended horizons, but differ in the primary source of difficulty: partial observability and error propagation in ALFWorld, compositional generalization in BabyAI-Text, and information-seeking under noise and cost constraints in WebShop. Together, these benchmarks provide a comprehensive testbed for evaluating long-horizon agent behavior under diverse failure modes.

\paragraph{Scope of the paper: Why not math or QA benchmarks}
We focus on interactive, long-horizon agent benchmarks where agents must act online and where latency and token-budget efficiency materially affect performance. Offline final-answer reasoning benchmarks such as math, code, and QA are out of scope, as they typically allow extended deliberation or backtracking without the compounding interaction delay central to our setting.

\subsection{Online Self-Improvement Protocol}
\label{app:online_protocol}

SAG updates the actor during a sequential online interaction stream. For each episode, the actor interacts with the environment using the current gate and critic. If the episode succeeds, we store the resulting context--action pairs $(x_t,a_t)$ in the buffer $\mathcal{D}$, where $a_t$ is the executed admissible action, either selected by the actor or suggested by the critic when critique is invoked. Although the actor prompt may elicit reasoning before action selection, online SFT is applied only to the executed action target, not to the generated reasoning trace.

\begin{figure}[h]
\centering
    \includegraphics[width=1\linewidth]{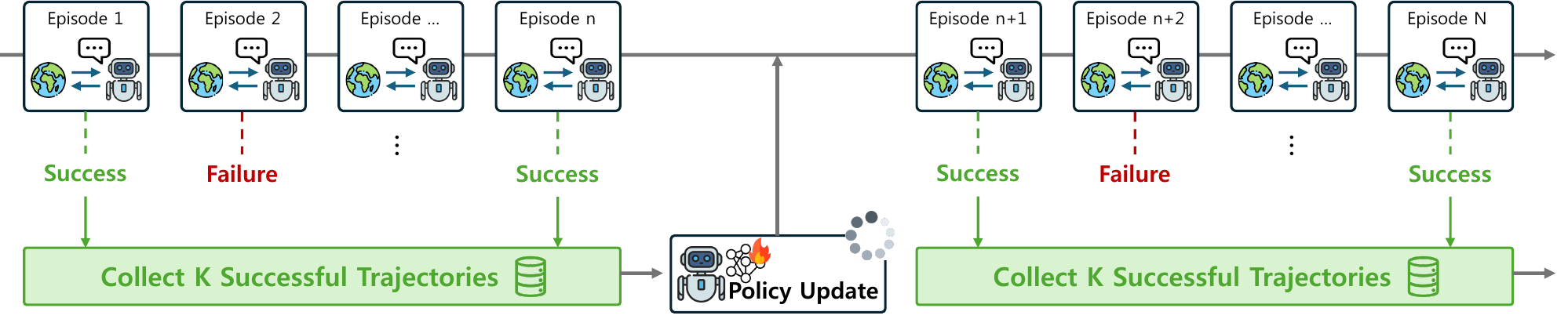}
    \vspace{-0.5cm}
    \caption{Overall online self-improvement process}
    \label{fig:onlinesft}
    \vspace{-0.4cm}
\end{figure}

Each episode is evaluated only once in the main protocol. If an episode later contributes successful context--action pairs to $\mathcal{D}$, it is not re-evaluated after the actor update; the updated actor is used only for subsequent episodes. After every $K$ successful episodes, we update only the actor with SFT on $\mathcal{D}$ while keeping the critic fixed. The overall online self-improvement process is provided in Figure~\ref{fig:onlinesft}.

At each update, we fine-tune from the same base actor rather than continuing from the previous updated actor. This follows bootstrapping-based self-training~\cite{zelikman2022star} and reduces drift from repeated updates. Unless otherwise stated, we use $K=10$ successful episodes per update. We study the sensitivity to $K$ in Appendix~\ref{app:update_sensitivity}.

\subsection{Hyperparameters}
\label{app:hyperparameter}
All of our experiments were conducted on a single NVIDIA RTX PRO 6000 Blackwell GPU with 96GB VRAM, and the detailed hyperparameters and training configurations for simulation are provided in Table~\ref{tab:sag_hparams}.

\begin{table}[h]
\centering
\small
\caption{SAG hyperparameters and online SFT configuration (shared across benchmarks unless otherwise noted).}
\begin{tabular}{l c l c}
\toprule
\textbf{Component} & \textbf{Value} & \textbf{Component} & \textbf{Value} \\
\midrule
\multicolumn{4}{l}{\textit{Uncertainty-based critique gate}} \\   
Gate threshold $\tau$        & 0.4 
& Margin threshold $\delta$             & 0.2 \\
\midrule
\multicolumn{4}{l}{\textit{Online SFT (self-improvement)}} \\ 
History window                         & 8 steps
& Max obs.\ length                     & 800 chars \\
Buffer capacity                        & 50{,}000
& Successes per update                 & 10 \\
Train batch size                       & 2
& Grad.\ accumulation                  & 8 \\
Train steps per update                 & 80
& Learning rate                        & $5\times10^{-5}$ \\
Max sequence length                    & 1024
& Optimizer                            & AdamW \\
Grad.\ clipping                        & 1.0
&  LoRA dropout                           & 0.05   \\
LoRA rank $r$                          & 8       
& LoRA scaling $\alpha$                & 16 \\

\bottomrule
\end{tabular}
\label{tab:sag_hparams}
\end{table}

\subsection{Detailed description of baselines}
\label{app:baseline}
We compare SAG against representative LLM agent frameworks.

\begin{itemize}[leftmargin=*, itemsep=0em, topsep=0em]
    \item \textbf{ReAct}~\cite{yao2022react}: A reasoning-and-acting framework that interleaves intermediate reasoning with environment actions. ReAct generates thoughts to decompose tasks and guide action selection while interacting with external environments. It serves as a baseline without explicit critique.
    \item \textbf{Reflexion}~\cite{single-reflexion}: A self-critique approach that generates reflections to guide future trials. Reflections are stored in memory and retrieved to improve subsequent attempts. Critique is applied in a fixed manner without adapting to step-level risk or cost.
    \item \textbf{Debate}~\cite{multi-debate}: A multi-agent deliberation approach where multiple agents discuss candidate solutions before acting. Structured argumentation helps identify errors and refine reasoning collaboratively. While improving robustness, it incurs high computational cost due to parallel agent executions.
    \item \textbf{CER}~\cite{feedback-CER}: A confidence-based framework that estimates uncertainty and uses it for selection or aggregation. It quantifies model confidence to filter low-quality outputs or combine predictions. While leveraging uncertainty signals, CER does not explicitly model critique as a decision nor invoke an external critic.
    \item \textbf{Traj-Boost}~\cite{ICL}: A trajectory-bootstrapping method that improves performance by reusing successful trajectories. It enhances learning from experience by reusing high-quality execution patterns. However, it does not provide execution-time critique or error correction during task execution.
    \item \textbf{Retrospex}~\cite{learning-retrospex}: An always-on critique framework that integrates a learned offline reinforcement learning critic by re-scoring candidate actions at every step. The critic estimates action quality based on historical trajectories, providing lightweight value assessments. Although the critic is lightweight, critique is applied throughout execution without selective gating.
    \item \textbf{LAC}~\cite{critic-LAC}: An always-on critique method that queries an LLM-based critic at every decision step. The critic provides numerical, value-based evaluations of candidate actions to improve action quality. The frequent critic invocation leads to substantial inference cost.
    \item \textbf{Ours (SAG)}: A cost-aware framework that treats critic invocation as a decision and selectively queries a critic based on actor ambiguity. By combining selective critique with online self-improvement, SAG aims to achieve strong performance with significantly improved token efficiency.
\end{itemize}
\subsection{Scope Relative to RL-based Agent Post-training}
\label{app:rl_scope}

SAG and RL-based agent post-training address complementary regimes of LLM agent improvement. 
RL-based methods such as PPO-style optimization, RLOO/REINFORCE-style optimization, GRPO, and GiGPO improve the underlying policy before deployment by collecting environment rollouts, computing rewards or advantages, and updating model parameters through policy optimization~\cite{ppo,grpo,gigpo}.
These methods can achieve strong task-specific performance when sufficient rollout budget, reward supervision, and optimization compute are available. 
SAG instead targets a deployment-time regime: given an available actor and a non-oracle critic, it asks when critique is worth its token and latency cost during interaction.

\begin{table}[h]
\centering
\small
\setlength{\tabcolsep}{5pt}
\renewcommand{\arraystretch}{1.15}
\caption{Scope comparison between RL-based agent post-training and SAG.}
\label{tab:rl_scope}
\resizebox{\linewidth}{!}{%
\begin{tabular}{lll}
\toprule
\textbf{Dimension} & \textbf{RL-based agent post-training} & \textbf{SAG} \\
\midrule
Primary goal 
& Improve the underlying policy before deployment 
& Decide when to use critique during deployment \\
Typical methods 
& PPO, GRPO, GiGPO 
& Gated critique + online SFT \\
Training signal 
& Rewards, preferences, or advantage estimates 
& Successful trajectories from online interaction \\
Rollout structure 
& Multiple sampled trajectories/completions per task 
& One interaction stream with selective critique \\
Models during training 
& Actor, often reference policy, optimizer states 
& Actor update; critic fixed and queried selectively \\
Cost focus 
& Training-time rollout and optimization budget 
& Inference-time token, critic-call, and latency cost \\
Hyperparameters 
& LR, KL, clipping, group size, reward normalization, rollout budget 
& Gate thresholds $\tau,\delta$ and update frequency \\
Comparison regime 
& Strong task-specific policy after post-training 
& Cost-aware improvement of an available actor \\
\bottomrule
\end{tabular}%
}
\end{table}

\paragraph{Different optimization stage}
RL-based post-training invests compute before deployment to produce a stronger policy. 
A typical RL pipeline repeatedly samples trajectories from the current policy, evaluates them with task rewards or preference-derived signals, estimates advantages, and performs policy updates, often with KL control against a reference policy. 
In contrast, SAG does not aim to replace this training stage. 
It operates during interaction, using a lightweight gate to decide whether an already available actor should query an auxiliary critic at the current step. 
Thus, RL methods primarily optimize \emph{what policy is learned}, whereas SAG optimizes \emph{when external critique should be used} under inference-time cost.

\paragraph{Different compute and memory regime}
The computational structure of RL-based post-training differs substantially from SAG. 
Group-based RL methods such as GRPO and GiGPO sample multiple trajectories or completions per task, compare them through reward or advantage estimates, and then reprocess these trajectories for policy-gradient updates. 
Many implementations also maintain a reference policy for KL regularization and store old log-probabilities, rollout buffers, and optimizer states. 
By contrast, SAG uses a single interaction trajectory at a time, invokes the critic only on selected ambiguous steps, and performs lightweight SFT on successful trajectories. 
In our implementation, only the actor is updated while the critic remains fixed. 
This makes SAG closer to inference-time adaptation or lightweight online SFT than to full RL post-training.

\paragraph{Different token-cost accounting}
RL-based methods may have low inference-time cost after training, but their training phase can consume large numbers of environment interactions and generated tokens due to repeated rollouts. 
This training-time token budget is not directly comparable to SAG's inference-time critique budget. 
SAG explicitly studies the cost incurred while the agent is solving tasks: token usage includes actor and critic tokens during interaction, and the method is designed to reduce this cost by avoiding unnecessary critique calls. 
A fair like-for-like comparison would therefore need to control not only final inference tokens, but also rollout budget, number of environment interactions, training tokens, GPU hours, reward supervision, and task-specific tuning.

\paragraph{Different practical assumptions}
RL-based post-training often requires task-specific reward design or reliable environment rewards, parallel rollout infrastructure, careful hyperparameter tuning, and repeated optimization over the target environment. 
For example, policy-gradient methods can be sensitive to learning rate, KL coefficient, clipping range, group size, reward normalization, and rollout budget. 
SAG assumes a different practical setting: the user already has an instruction-tuned actor and an imperfect critic, and wants to improve reliability during deployment without large-scale task-specific RL training. 
Its main hyperparameters are the gate threshold $\tau$ and margin threshold $\delta$, which we find to be robust across a broad range in Appendix~\ref{app:threshold_sensitivity}.

\paragraph{Complementarity}
These differences do not imply that RL-based post-training and SAG are mutually exclusive. 
A stronger RL-trained agent could still benefit from SAG at deployment time, since even strong policies may encounter ambiguous states where external critique is useful. 
Conversely, SAG can provide a lightweight alternative when large-scale RL training is impractical or when the goal is to manage token and latency costs during online interaction. 
We therefore treat RL-based agent post-training as complementary to SAG, and focus our main empirical comparisons on inference-time and lightweight-adaptation baselines that share SAG's deployment-time regime.

\section{Extra Simulation Results}
\label{app:result}

\subsection{Extended Model Scaling with Always-on Critique}
\label{app:extended_scaling}

We extend the capacity analysis to larger Qwen backbones up to 32B and compare three configurations: 
\textbf{w/o critic}, which uses the actor without critique; 
\textbf{always-on critic}, which removes the gate and invokes the critic at every step; and 
\textbf{SAG}, which uses gated critique. 
For SAG, the actor and critic use the same backbone and model size. We report success rate, token usage, and critic calls per episode. 
Across benchmarks and model sizes, SAG consistently improves over the no-critic baseline and usually matches or exceeds always-on critique while using far fewer critic calls and substantially lower token cost.

\begin{table}[h]
\centering
\small
\setlength{\tabcolsep}{3.5pt}
\renewcommand{\arraystretch}{1.12}
\caption{Extended scaling results with no critic, always-on critic, and SAG. Token usage includes actor and critic tokens.}
\label{tab:extended_scaling}
\begin{tabular}{llccccccccc}
\toprule
\multirow{2}{*}{\textbf{Benchmark}} 
& \multirow{2}{*}{\textbf{Backbone}}
& \multicolumn{3}{c}{\textbf{w/o critic}}
& \multicolumn{3}{c}{\textbf{always-on critic}}
& \multicolumn{3}{c}{\textbf{SAG}} \\
\cmidrule(lr){3-5}\cmidrule(lr){6-8}\cmidrule(lr){9-11}
& & Succ. & Token & Call & Succ. & Token & Call & Succ. & Token & Call \\
\midrule
\multirow{4}{*}{ALFWorld}
& 3B  & 21.6 & 57 & -- & 59.0 & 271 & 15.1 & \textbf{69.4} & 58 & 3.2 \\
& 7B  & 25.4 & 55 & -- & 74.8 & 273 & 14.7 & \textbf{78.3} & 56 & 2.6 \\
& 14B & 77.2 & 49 & -- & 80.5 & 292 & 13.3 & \textbf{82.1} & 53 & 2.3 \\
& 32B & 79.1 & 48 & -- & 86.2 & 308 & 13.2 & \textbf{87.2} & 54 & 2.1 \\
\midrule
\multirow{4}{*}{BabyAI}
& 3B  & 34.0 & 59 & -- & 52.3 & 356 & 10.3 & \textbf{61.8} & 79 & 3.3 \\
& 7B  & 38.1 & 55 & -- & 69.6 & 377 & 10.1 & \textbf{70.0} & 78 & 2.8 \\
& 14B & 53.3 & 53 & -- & 76.2 & 382 & 8.5  & \textbf{76.3} & 71 & 2.6 \\
& 32B & 61.7 & 53 & -- & \textbf{86.2} & 308 & 8.4 & 82.8 & 68 & 2.3 \\
\midrule
\multirow{4}{*}{WebShop}
& 3B  & 10.9 & 11 & -- & 25.7 & 54 & 6.8 & \textbf{34.2} & 11 & 2.3 \\
& 7B  & 13.2 & 11 & -- & 33.2 & 57 & 6.5 & \textbf{37.0} & 9 & 2.2 \\
& 14B & 18.9 & 12 & -- & 41.0 & 61 & 5.9 & \textbf{45.7} & 9 & 1.8 \\
& 32B & 25.3 & 9  & -- & 43.5 & 62 & 5.7 & \textbf{48.1} & 9 & 1.6 \\
\bottomrule
\end{tabular}%
\end{table}

Two trends are consistent across benchmarks. First, selective critique remains useful even for stronger actors: SAG improves over the no-critic baseline at every model scale. Second, always-on critique often improves success but incurs much higher token cost and critic-call frequency. SAG achieves a better performance--cost trade-off by concentrating critique on ambiguous steps rather than invoking it throughout the trajectory.
\subsection{Effect of Critic Capacity}
\label{app:critic_capacity}

We analyze whether SAG benefits from stronger critics by fixing the actor to Qwen2.5-7B-Instruct and varying the critic size on ALFWorld.

\begin{table}[h]
\centering
\small
\setlength{\tabcolsep}{8pt}
\renewcommand{\arraystretch}{1.15}
\caption{Effect of critic capacity with a fixed 7B actor on ALFWorld.}
\label{tab:critic_capacity}
\begin{tabular}{lc}
\toprule
\textbf{Actor / Critic} & \textbf{Succ.} $\uparrow$ \\
\midrule
7B / No critic & 25.4 \\
7B / 3B critic & 76.9 \\
7B / 7B critic & 78.3 \\
7B / 14B critic & \textbf{82.1} \\
\bottomrule
\end{tabular}
\end{table}

A stronger critic further improves success, but even a smaller 3B critic provides a large gain over the no-critic baseline. This suggests that SAG does not require an oracle or a substantially stronger model; rather, it benefits from selectively invoking an auxiliary perspective at ambiguous decision points.

\subsection{Matched Online-SFT Controls}
\label{app:self_improve_controls}

To isolate the effect of selective critique from online self-improvement alone, we compare SAG with two matched online-SFT controls. 
\textbf{ReAct + online SFT} uses the same successful-trajectory collection and actor-update protocol as SAG, but does not use any critic. 
\textbf{LAC + online SFT} uses the same update protocol with always-on critique, invoking the critic at every step. 
Thus, all three methods share the same online self-improvement mechanism, but differ in how critique is used.

\begin{table}[h]
\centering
\small
\setlength{\tabcolsep}{5pt}
\renewcommand{\arraystretch}{1.15}
\caption{Matched online-SFT controls across all benchmarks. All methods use the same online update protocol. Token usage includes both actor and critic tokens.}
\label{tab:matched_sft_full}
\begin{tabular}{lcccccc}
\toprule
\multirow{2}{*}{\textbf{Method}} 
& \multicolumn{2}{c}{\textbf{ALFWorld}} 
& \multicolumn{2}{c}{\textbf{BabyAI-Text}} 
& \multicolumn{2}{c}{\textbf{WebShop}} \\
\cmidrule(lr){2-3}\cmidrule(lr){4-5}\cmidrule(lr){6-7}
& \textbf{Succ.} $\uparrow$ & \textbf{\#Token} $\downarrow$
& \textbf{Succ.} $\uparrow$ & \textbf{\#Token} $\downarrow$
& \textbf{Succ.} $\uparrow$ & \textbf{\#Token} $\downarrow$ \\
\midrule
ReAct + online SFT 
& 30.2 & 57 
& 40.8 & 56 
& 15.3 & 12 \\
LAC + online SFT 
& 74.2 & 341 
& 67.2 & 438 
& 32.1 & 84 \\
\rowcolor{oursgray}
\textbf{SAG} 
& \textbf{78.3} & \textbf{56} 
& \textbf{70.0} & \textbf{78} 
& \textbf{37.0} & \textbf{9} \\
\bottomrule
\end{tabular}%

\end{table}

The results show that online self-improvement alone does not explain SAG's gains. 
ReAct + online SFT improves only modestly because it can learn only from trajectories generated without critique, which limits the quality and diversity of successful supervision. 
LAC + online SFT benefits from critic-assisted trajectories, but its always-on critique incurs substantially higher token cost: $341$k on ALFWorld, $438$k on BabyAI-Text, and $84$k on WebShop. 
SAG achieves the best success on all three benchmarks while using token budgets close to or lower than ReAct + online SFT and far below LAC + online SFT. 
This confirms that the improvement comes from the combination of selective critique and online self-improvement, rather than from online SFT alone.

\subsection{Robustness to Trajectory Ordering}
\label{app: shuffled_order}
To verify the robustness of the online self-improvement method, we repeated the ALFWorld online evaluation under different random episode orders. As online SFT updates depend on the accumulation of successful trajectories, episode order could potentially influence when updates occur and how much improvement is observed. We therefore compare the default benchmark order with shuffled orders using different seeds. 

\begin{table}[h]
    \centering
    \caption{Robustness of online-SFT across different episode orders}
    \begin{tabular}{lccc}
    \toprule
         \textbf{Order Setting} & \textbf{Seed / Order} & \textbf{Success (\%)} & \textbf{Token (k)} \\
    \midrule
        Default benchmark order & default & $78.3$ & $56$ \\
        Shuffled order & 1 & $78.7$ & $56$ \\
        Shuffled order & 2 & $78.2$ & $58$ \\
        \rowcolor{oursgray}
        Mean $\pm$ std over shuffled orders & default/1/2 & 78.3 $\pm$ 0.3 & 56.7 $\pm$ 1.2 \\
    \bottomrule
    \end{tabular}
    \label{tab:error_bars}
\end{table}

The results show that success rate and token usage remain stable across order settings, and the qualitative conclusion remains unchanged. This suggests that the gains are not driven by one favorable sequence of episodes, but are robust to task order. 

\subsection{Gate Threshold Sensitivity}
\label{app:threshold_sensitivity}

SAG uses the same thresholds across benchmarks and backbones unless otherwise specified. We analyze the sensitivity to the entropy threshold $\tau$ and margin threshold $\delta$ on ALFWorld. The results show that $\tau$ primarily controls the cost--success trade-off, while $\delta$ has a relatively mild effect on success over a broad range.

\begin{table}[h]
\centering
\small
\setlength{\tabcolsep}{8pt}
\renewcommand{\arraystretch}{1.15}
\caption{Sensitivity to $\tau$ and $\delta$ on ALFWorld.}
\label{tab:threshold_sensitivity}
\begin{tabular}{cccccc}
\toprule
$\tau$ & \textbf{Succ.} $\uparrow$ & \textbf{\#Token} $\downarrow$
& $\delta$ & \textbf{Succ.} $\uparrow$ & \textbf{\#Token} $\downarrow$ \\
\midrule
0.2 & 79.0 & 74 & 0.1 & 77.1 & 53 \\
0.3 & 78.9 & 63 & 0.2 & \textbf{78.3} & 56 \\
0.4 & \textbf{78.3} & 56 & 0.4 & 78.1 & 61 \\
0.5 & 75.1 & 49 & 0.5 & 77.8 & 64 \\
0.7 & 66.1 & 45 & 1.0 & 75.3 & 60 \\
\bottomrule
\end{tabular}
\end{table}

Performance degrades mainly when $\tau$ is set too high, causing the agent to under-use critique. In contrast, nonzero values of $\delta$ yield similar success, indicating that the local ambiguity term is useful but does not require fine-grained tuning.

\newpage
\subsection{Simulation results on BabyAI-Text}
\label{app:baby}
\paragraph{Parameter sensitivity}
\begin{figure}[hbt!] 
    \centering
    \begin{subfigure}[t]{\linewidth}
        \centering
        \includegraphics[width=0.5\linewidth]{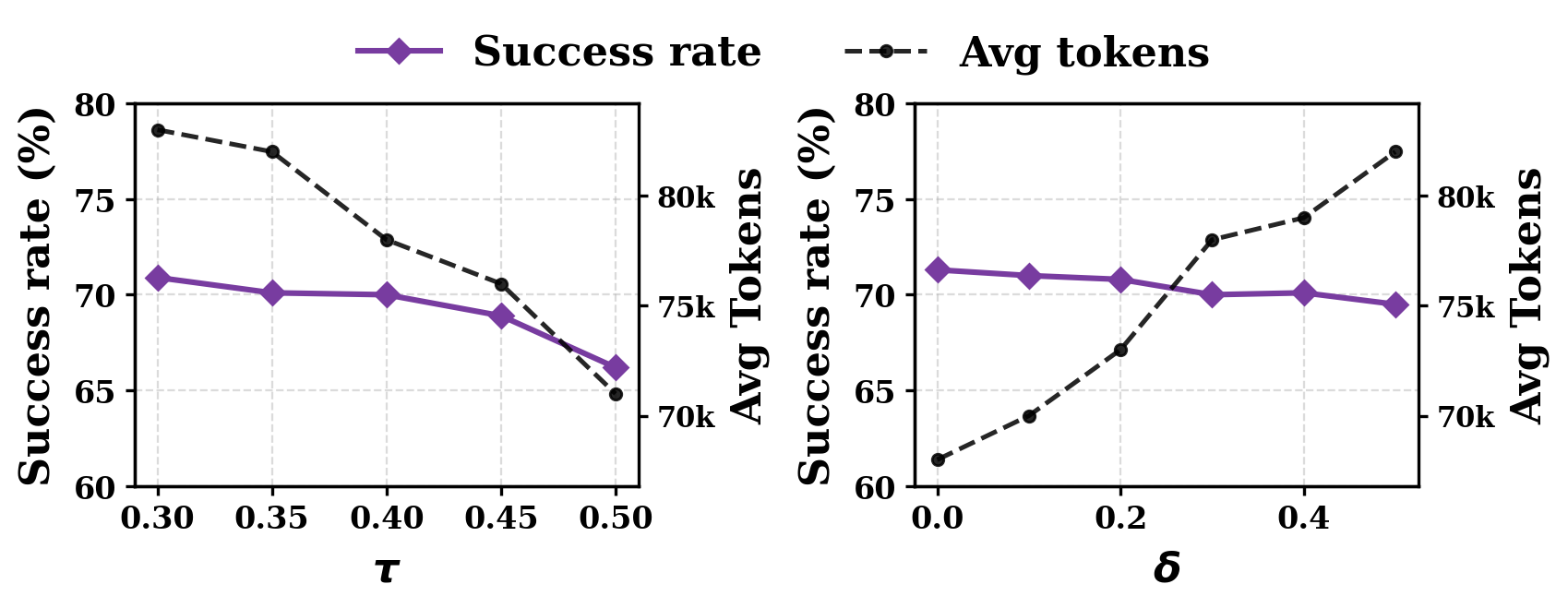}
    \end{subfigure}
    \vspace{-0.6cm}
    \caption{Parameter sensitivity analysis on BabyAI-Text benchmark}
    \label{fig:sensitivity_baby}
    \vspace{-0.3cm}
\end{figure}
Figure~\ref{fig:sensitivity_baby} reports the sensitivity of SAG to the gate threshold $\tau$ and the margin threshold $\delta$ on BabyAI-Text.
Varying $\tau$ yields a smooth cost--performance trade-off: as $\tau$ increases from $0.30$ to $0.50$, success decreases moderately from $70.9\%$ to $66.2\%$, while the average token usage drops substantially from $83$k to $71$k tokens.
Within the mid-range ($\tau \in [0.35, 0.45]$), success remains relatively stable around $70\%$, but token cost continues to decrease ($82$k $\rightarrow 76$k), indicating that $\tau$ can be used as an effective knob for cost calibration.
In contrast, SAG is largely insensitive to the margin threshold $\delta$: success stays within a narrow band of $69.5\%$--$71.3\%$ across all tested values, while larger $\delta$ primarily increases token usage from $68$k to $82$k by triggering critique more often.
These trends mirror those observed on ALFWorld, suggesting that $\tau$ governs the main cost--performance trade-off, whereas $\delta$ can be chosen from a broad range without fine-grained tuning.

\begin{figure*}[h] 
    \centering
    \begin{subfigure}[t]{0.24\linewidth}
        \centering
        \includegraphics[width=\linewidth]{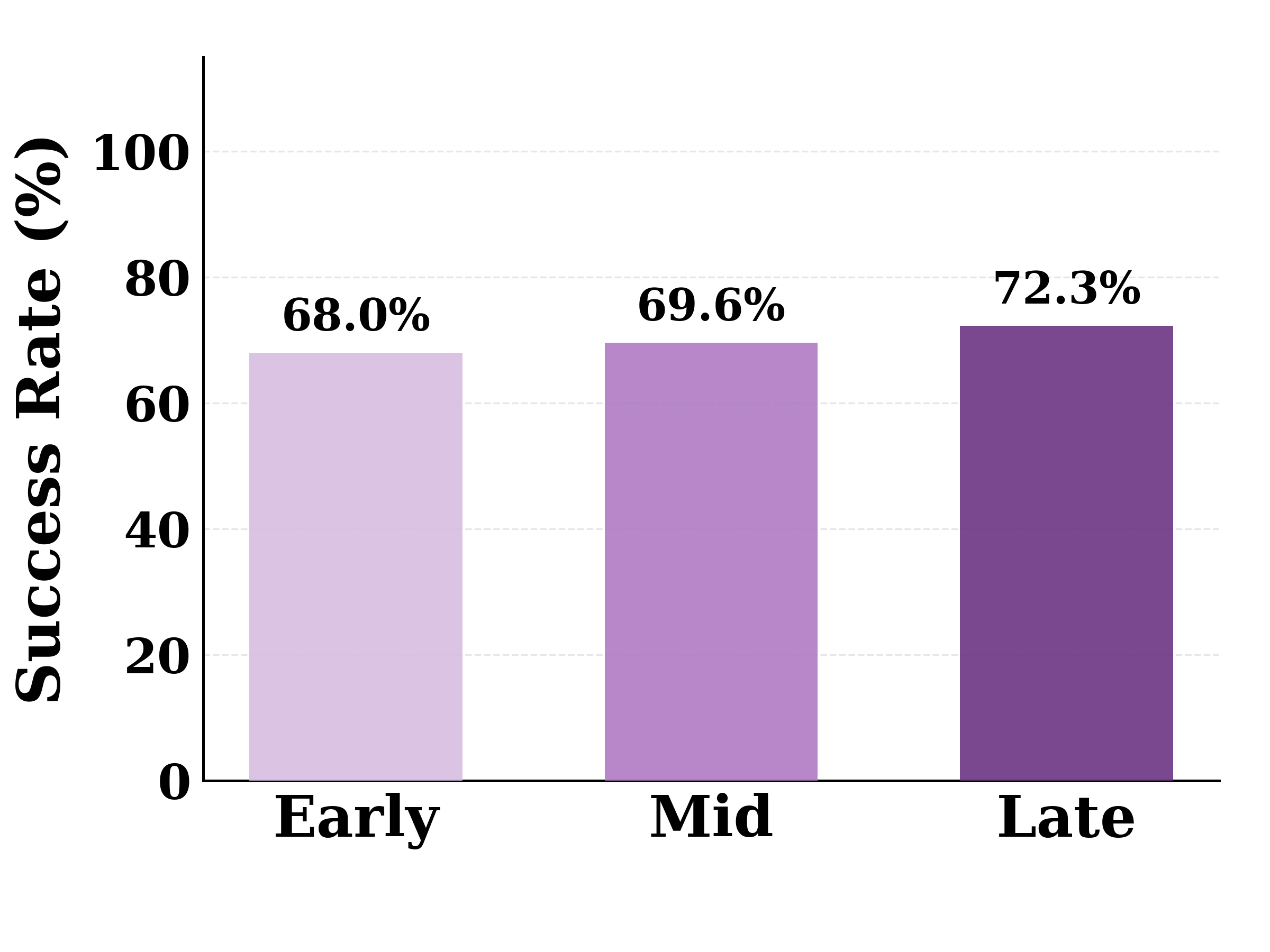}
        \vspace{-0.7cm}
        \caption{Success rate (\%)}
        \label{fig:learn_succ_baby}
    \end{subfigure}
    \hfill
    \begin{subfigure}[t]{0.24\linewidth}
        \centering
        \includegraphics[width=\linewidth]{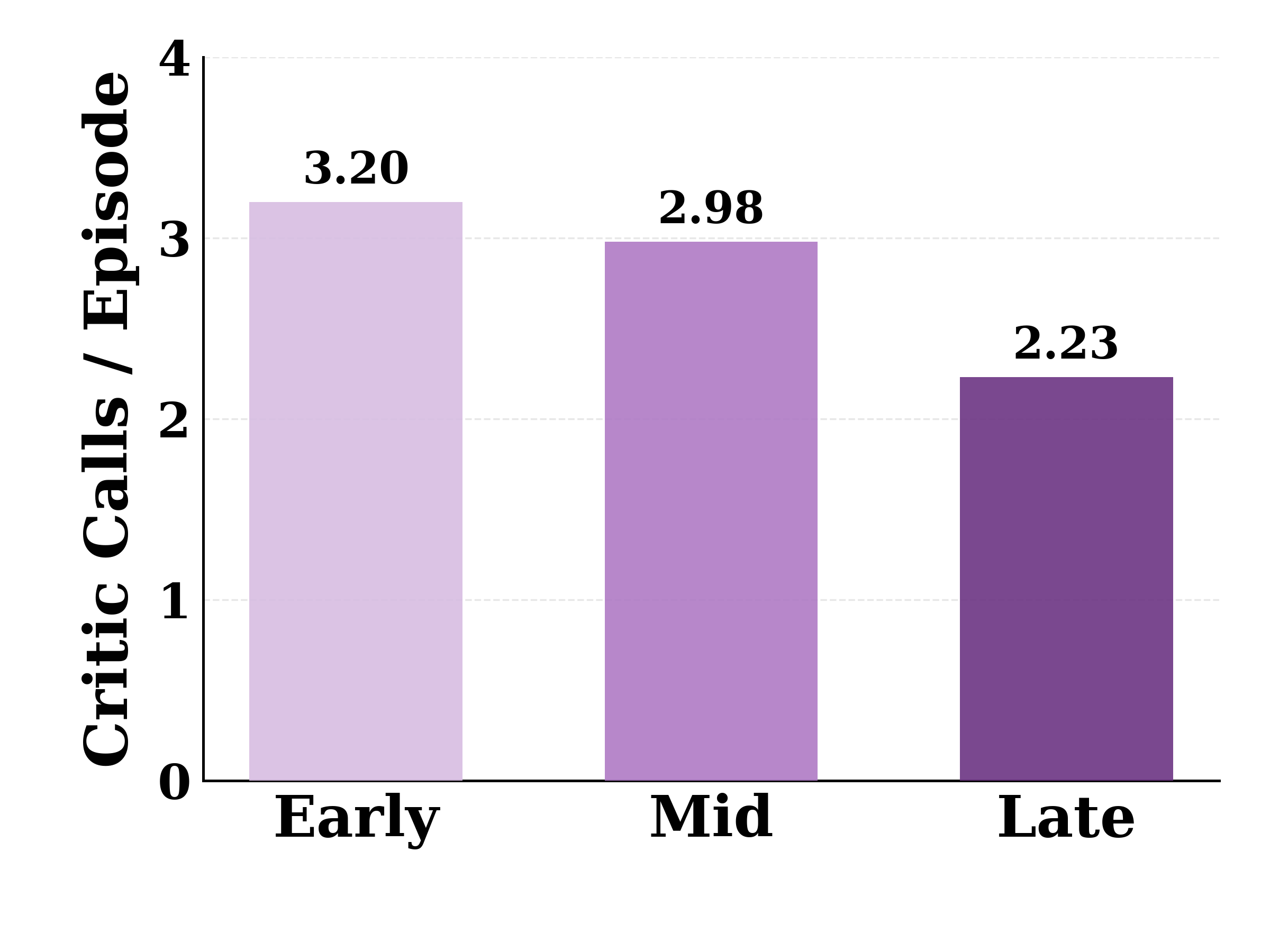}
        \vspace{-0.7cm}
        \caption{Critic calls / episode}
        \label{fig:learn_critic_baby}
    \end{subfigure}
    \hfill
    \begin{subfigure}[t]{0.24\linewidth}
        \centering
        \includegraphics[width=\linewidth]{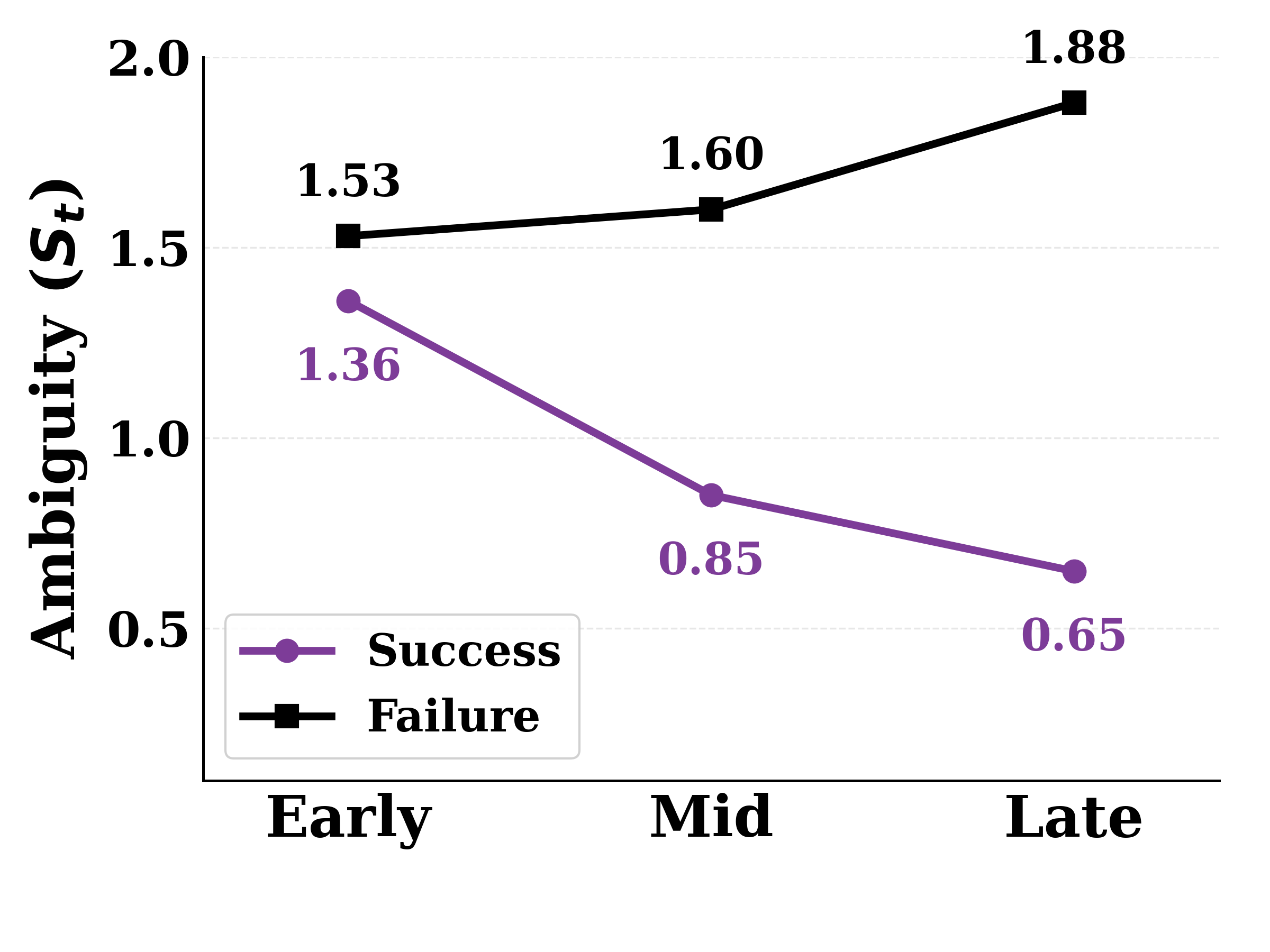}
        \vspace{-0.7cm}
        \caption{$S_t$ for success vs.\ failure}
        \label{fig:learn_succfail_baby}
    \end{subfigure}
    \hfill
    \begin{subfigure}[t]{0.24\linewidth}
        \centering
        \includegraphics[width=\linewidth]{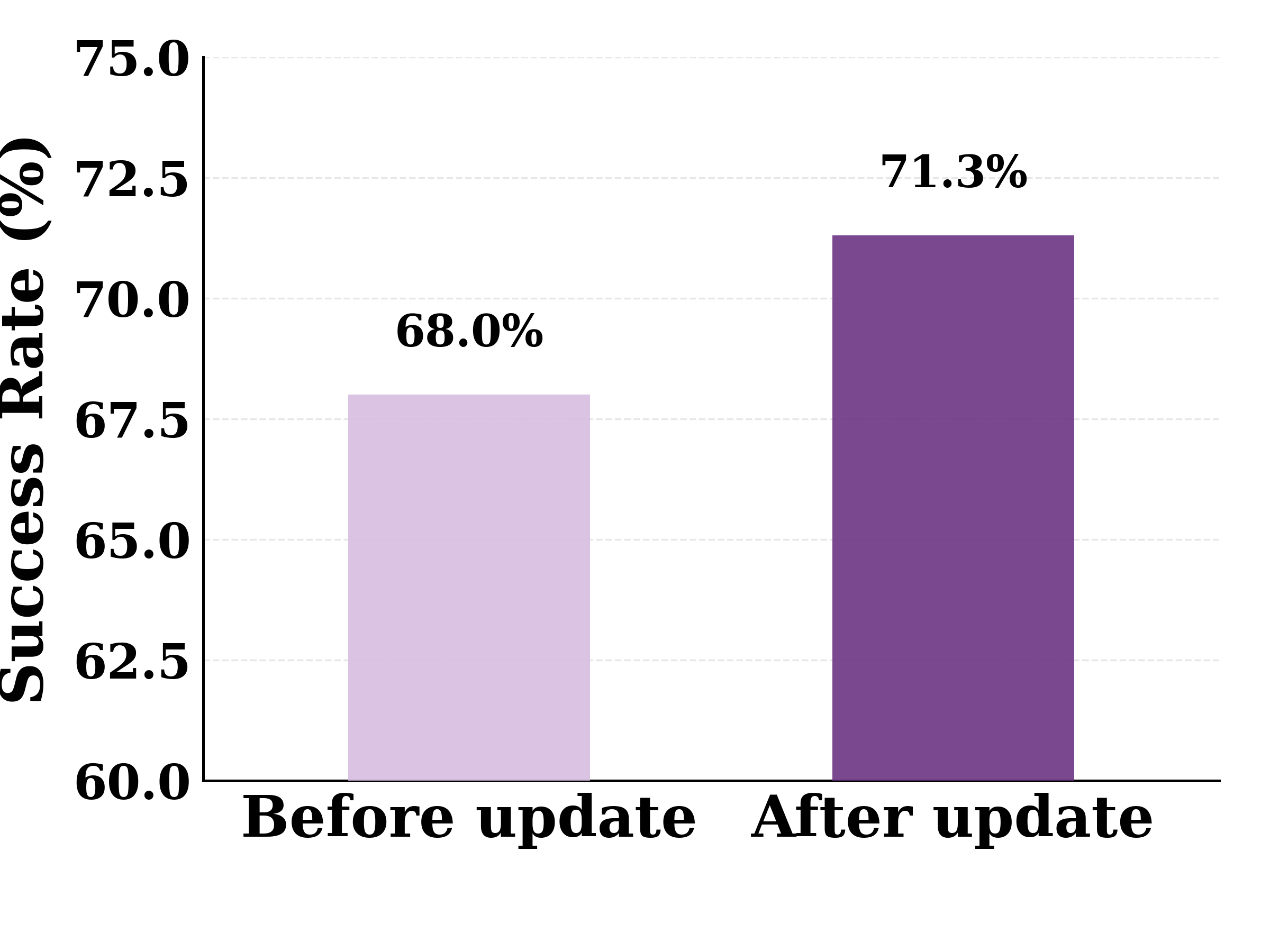}
        \vspace{-0.7cm}
        \caption{Seen: before vs.\ after}
        \label{fig:learn_beforeafter_baby}
    \end{subfigure}
   \vspace{-0.1cm}
    \caption{Learning dynamics of SAG under online self-improvement. Metrics are aggregated into Early/Mid/Late phases over 100 evaluation episodes on BabyAI-Text benchmark.}
    \label{fig:learning_dynamics_baby}
    \vspace{-0.3cm}
\end{figure*}

\paragraph{Learning dynamics under online self-improvement}
Figure~\ref{fig:learning_dynamics_baby} shows the learning dynamics of SAG under online self-improvement on the BabyAI-Text benchmark.
As interaction proceeds, the success rate steadily improves from $68.0\%$ (Early) to $69.6\%$ (Mid) and $72.3\%$ (Late; Figure~\ref{fig:learn_succ_baby}), while the average number of critic calls per episode decreases from $3.20$ to $2.98$ and finally $2.23$ (Figure~\ref{fig:learn_critic_baby}).
This reflects SAG's intended behavior: as successful trajectories are distilled into the actor via online SFT, the actor increasingly resolves decisions on its own under the same gating rule.

Figure~\ref{fig:learn_succfail_baby} further shows that ambiguity remains well separated between successes and failures over time.
For successful episodes, the ambiguity score $S_t$ decreases from $1.36$ to $0.65$, whereas failures maintain higher ambiguity and even become more uncertain ($1.53 \rightarrow 1.88$), indicating that remaining errors concentrate on harder decision points.
Moreover, performance on previously seen episodes improves after online updates, increasing from $68.0\%$ to $71.3\%$ (Figure~\ref{fig:learn_beforeafter_baby}), suggesting that SAG does not forget past experience and can correct earlier failures.
Overall, these results corroborate the findings on ALFWorld: critique acts as a transient scaffold, with SAG gradually internalizing effective behaviors and shifting decisions toward lower-risk states without increasing inference-time computation.

\subsection{Simulation results on WebShop}
\label{app:web}

\paragraph{Parameter sensitivity}
\begin{figure}[hbt!] 
    \centering
    \begin{subfigure}[t]{\linewidth}
        \centering
        \includegraphics[width=0.5\linewidth]{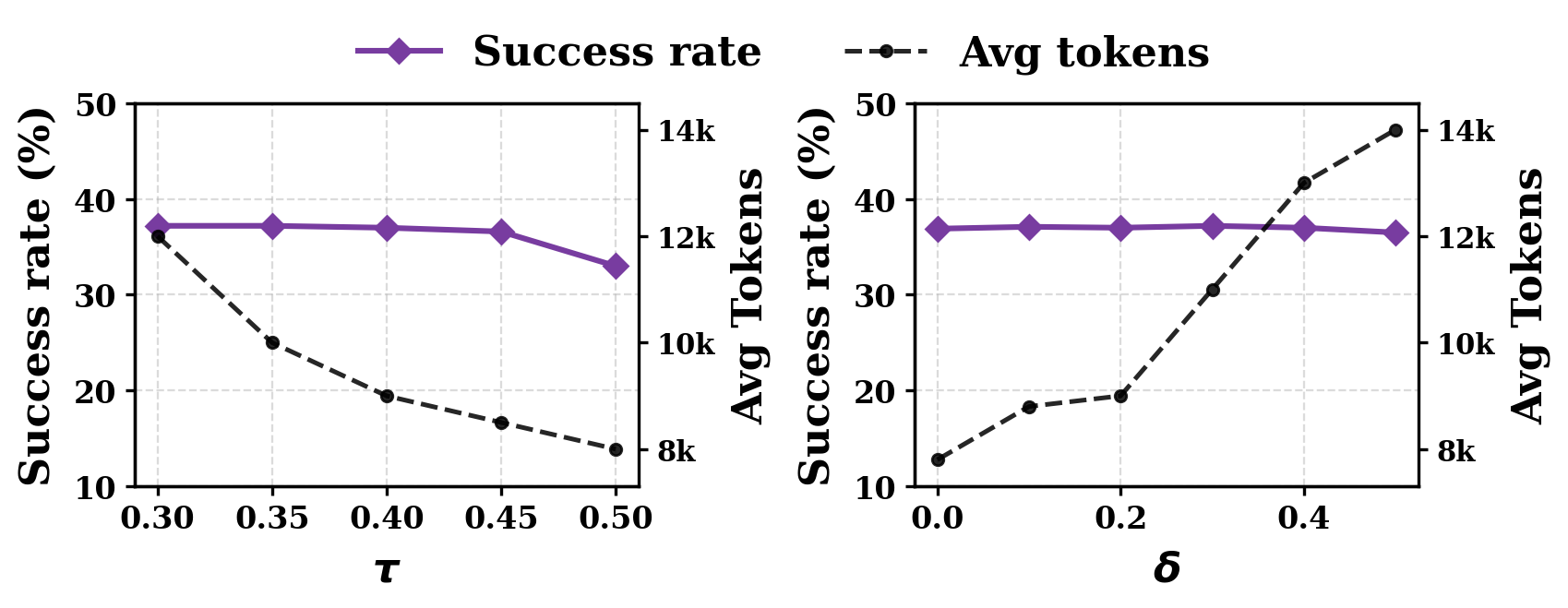}
    \end{subfigure}
    \vspace{-0.6cm}
    \caption{Parameter sensitivity analysis on WebShop benchmark}
    \label{fig:sensitivity_web}
    \vspace{-0.3cm}
\end{figure}

Figure~\ref{fig:sensitivity_web} analyzes the sensitivity of SAG to $\tau$ and $\delta$ on the WebShop benchmark.
Increasing $\tau$ leads to a monotonic reduction in token usage ($12$k $\rightarrow 8$k tokens), while success decreases from $37.2\%$ to $33.0\%$.
This reflects the higher sensitivity of WebShop to missed critique in open-domain information-seeking tasks.
By contrast, success remains stable across a wide range of $\delta$ values ($36.9\%$--$37.2\%$), indicating robustness to the local ambiguity threshold.
As in other environments, larger $\delta$ increases critique frequency and token usage, confirming that $\delta$ primarily affects cost rather than reliability.

\begin{figure*}[h] 
    \centering
    \begin{subfigure}[t]{0.24\linewidth}
        \centering
        \includegraphics[width=\linewidth]{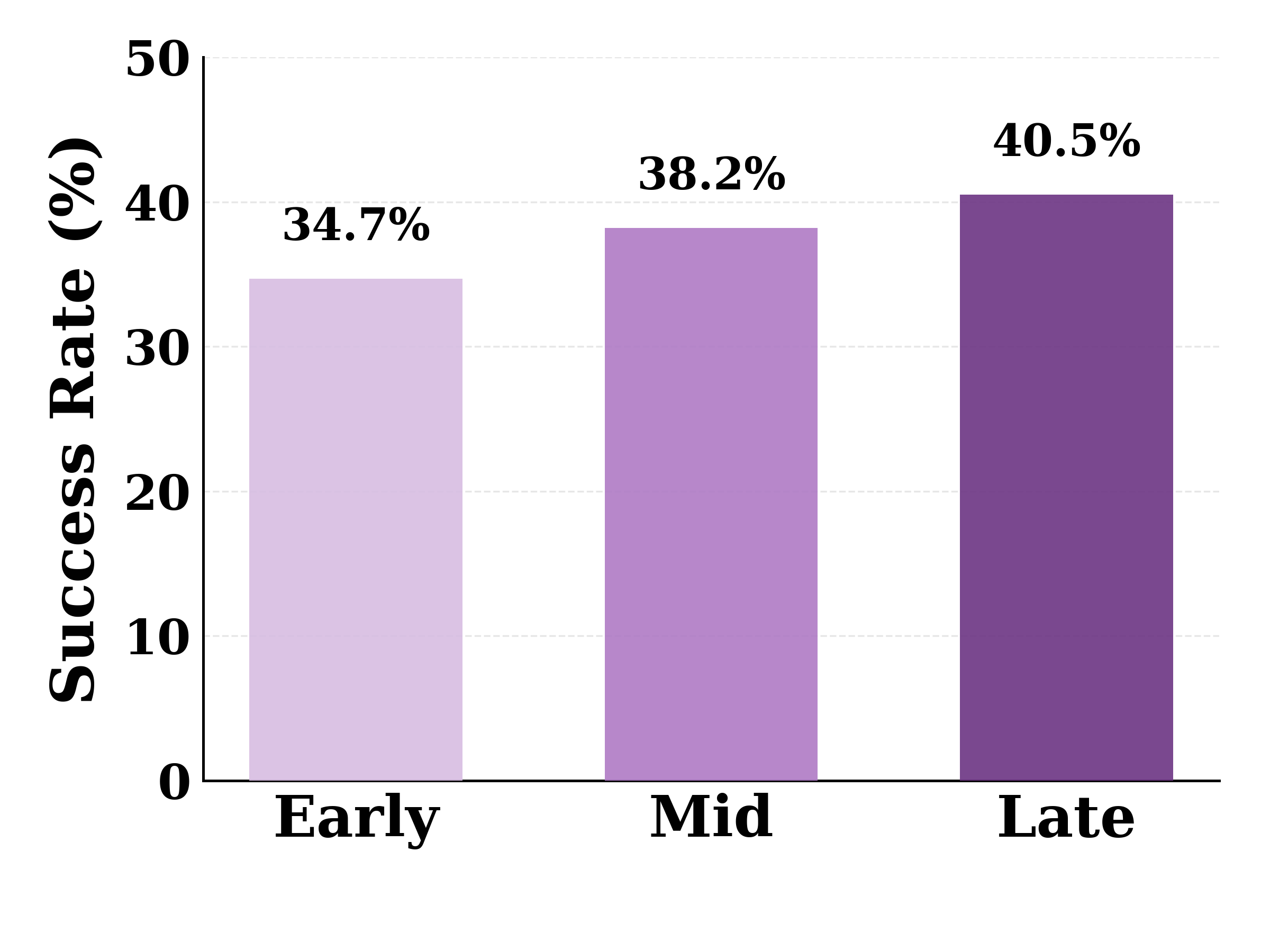}
        \vspace{-0.7cm}
        \caption{Success rate (\%)}
        \label{fig:learn_succ_web}
    \end{subfigure}
    \hfill
    \begin{subfigure}[t]{0.24\linewidth}
        \centering
        \includegraphics[width=\linewidth]{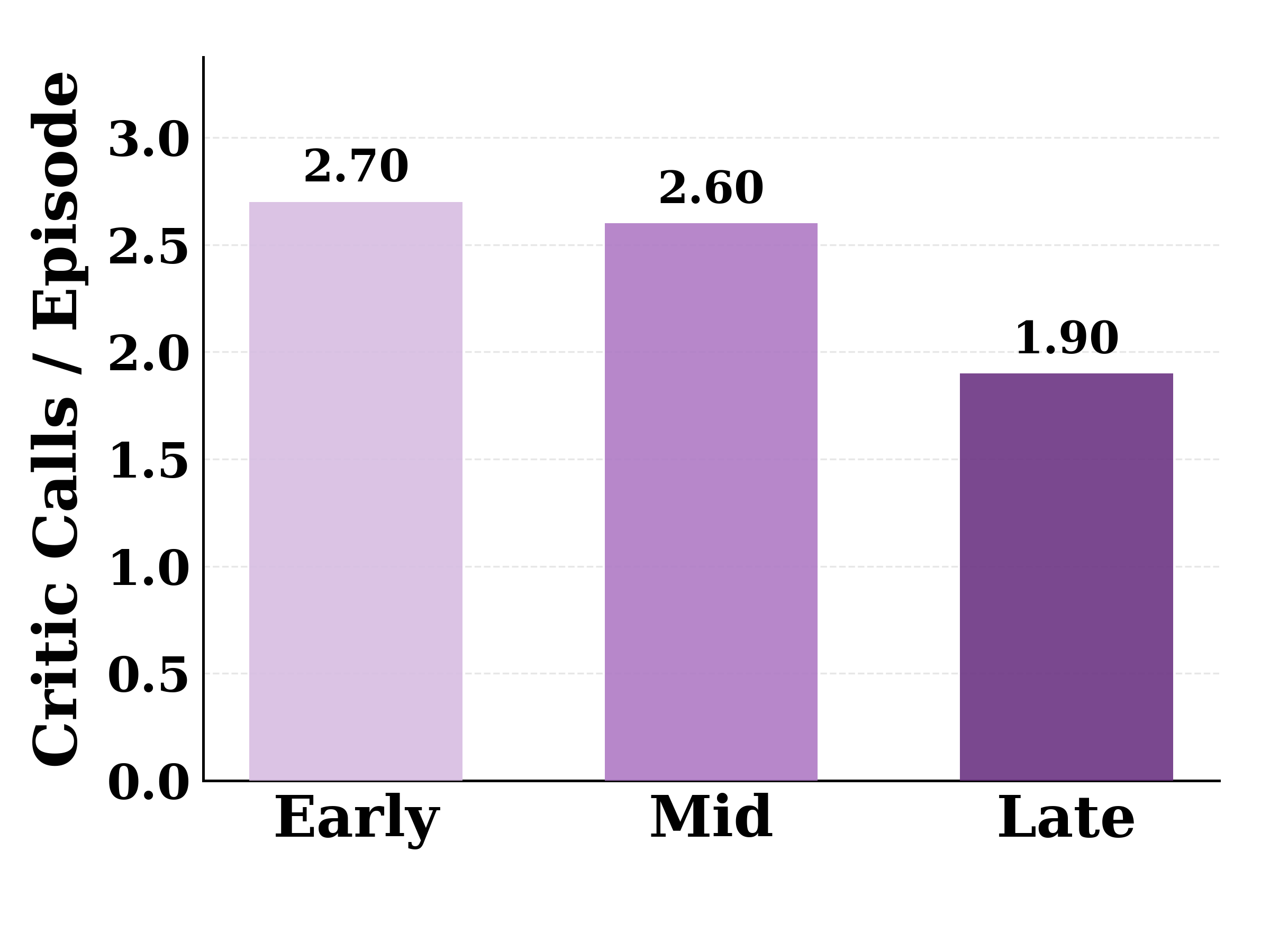}
        \vspace{-0.7cm}
        \caption{Critic calls / episode}
        \label{fig:learn_critic_web}
    \end{subfigure}
    \hfill
    \begin{subfigure}[t]{0.24\linewidth}
        \centering
        \includegraphics[width=\linewidth]{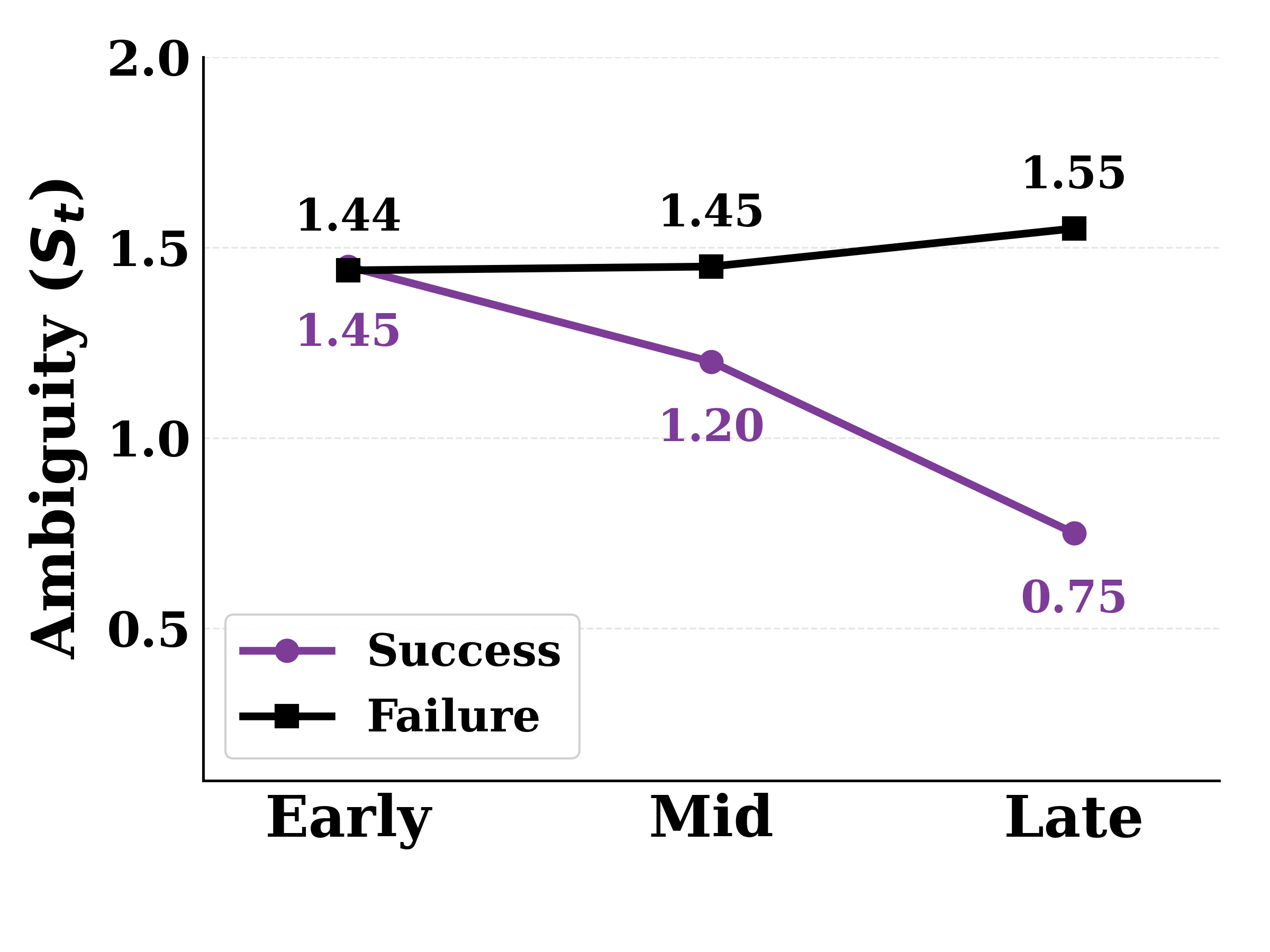}
        \vspace{-0.7cm}
        \caption{$S_t$ for success vs.\ failure}
        \label{fig:learn_succfail_web}
    \end{subfigure}
    \hfill
    \begin{subfigure}[t]{0.24\linewidth}
        \centering
        \includegraphics[width=\linewidth]{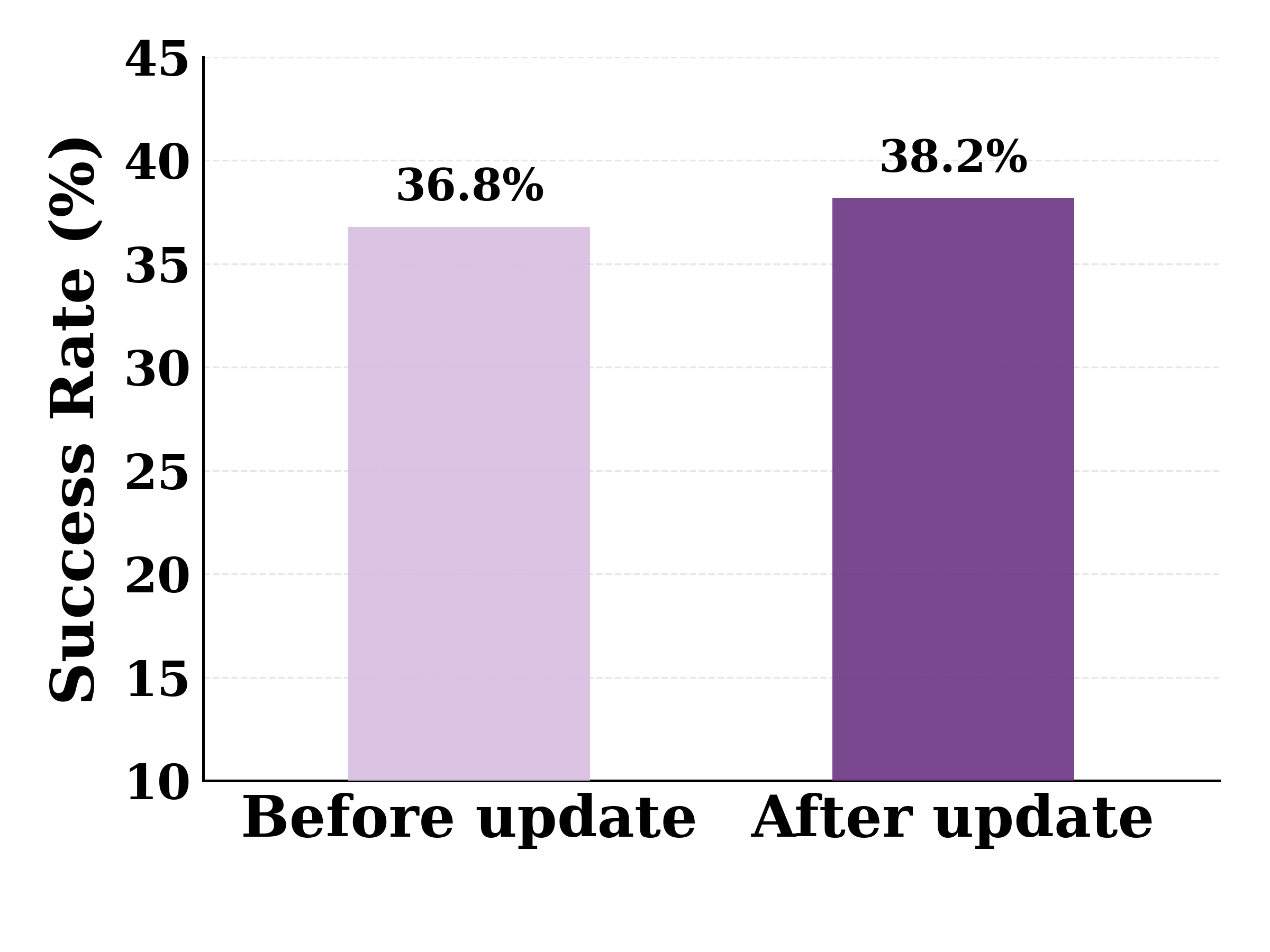}
        \vspace{-0.7cm}
        \caption{Seen: before vs.\ after}
        \label{fig:learn_beforeafter_web}
    \end{subfigure}
   \vspace{-0.1cm}
    \caption{Learning dynamics of SAG under online self-improvement. Metrics are aggregated into Early/Mid/Late phases over 200 evaluation episodes on WebShop benchmark.}
    \label{fig:learning_dynamics_web}
    \vspace{-0.3cm}
\end{figure*}

\paragraph{Learning dynamics under online self-improvement}
Figure~\ref{fig:learning_dynamics_web} presents the learning dynamics of SAG on WebShop.
Success rate steadily improves from $34.7\%$ to $38.2\%$ and $40.5\%$ over the Early/Mid/Late phases (Figure~\ref{fig:learn_succ_web}), while the average number of critic calls per episode decreases from $2.7$ to $1.9$ (Figure~\ref{fig:learn_critic_web}).
Ambiguity remains well separated between successes and failures throughout learning (Figure~\ref{fig:learn_succfail_web}): successful trajectories exhibit decreasing ambiguity ($1.45 \rightarrow 0.75$), whereas failures remain consistently more ambiguous.
Moreover, performance on previously seen episodes improves after online updates, increasing from $36.8\%$ to $38.2\%$ (Figure~\ref{fig:learn_beforeafter_web}).
These results align with those observed on ALFWorld and BabyAI-Text, demonstrating that SAG consistently transforms critique from a frequent intervention into a transient scaffold that accelerates learning while preserving inference efficiency.

\subsection{Online SFT vs. One-shot SFT}
\label{app:oneshot_sft}

We compare SAG's online self-improvement with a one-shot SFT variant that first collects successful trajectories and then performs a single actor update. Online-SFT incorporates critic-corrected behaviors earlier, allowing subsequent rollouts to benefit from the updated actor and reducing later reliance on critique.

\begin{table}[h]
\centering
\small
\setlength{\tabcolsep}{5pt}
\renewcommand{\arraystretch}{1.15}
\caption{Comparison between one-shot SFT and online-SFT.}
\label{tab:oneshot_online_sft}
\begin{tabular}{lcccccc}
\toprule
\multirow{2}{*}{\textbf{Method}} 
& \multicolumn{2}{c}{\textbf{ALFWorld}} 
& \multicolumn{2}{c}{\textbf{BabyAI-Text}} 
& \multicolumn{2}{c}{\textbf{WebShop}} \\
\cmidrule(lr){2-3}\cmidrule(lr){4-5}\cmidrule(lr){6-7}
& \textbf{Succ.} $\uparrow$ & \textbf{\#Token} $\downarrow$
& \textbf{Succ.} $\uparrow$ & \textbf{\#Token} $\downarrow$
& \textbf{Succ.} $\uparrow$ & \textbf{\#Token} $\downarrow$ \\
\midrule
One-shot SFT 
& 72.8 & 63 
& 66.9 & 82 
& 34.8 & 10 \\
\rowcolor{oursgray}
Online SFT (SAG) 
& \textbf{78.3} & \textbf{56} 
& \textbf{70.0} & \textbf{78} 
& \textbf{37.0} & \textbf{9} \\
\bottomrule
\end{tabular}%
\end{table}

Online SFT consistently improves success while maintaining comparable or lower token usage. This supports the view that early distillation of critic-assisted successful behavior improves subsequent interaction, whereas one-shot SFT misses this compounding benefit.

\newpage
\subsection{Failure Modes and Non-oracle Critic Behavior}
\label{app:failure_cases}

SAG assumes that critique is useful but not perfect. We observe three common failure modes.

\paragraph{Incorrect critic suggestion}
The critic may suggest an action that is plausible but suboptimal under the current environmental state. This can occur when the critic misinterprets the interaction history or overlooks a constraint in the admissible action set.

\paragraph{Missed critique opportunity}
The gate may fail to invoke critique when the actor is confidently wrong. This false-negative case is difficult for any actor-side uncertainty method, since the actor's distribution does not expose the underlying error.

\paragraph{Unrecoverable early error}
In long-horizon environments, some early mistakes create states from which later critique cannot recover. This suggests that future work could combine step-level gating with trajectory-level monitoring.

These cases highlight that SAG does not rely on an oracle critic. Its gains come from invoking imperfect critique selectively at high-risk steps, while avoiding the systematic cost of always-on feedback.

\subsection{Sensitivity to Online Update Frequency}
\label{app:update_sensitivity}

We vary the number of successful episodes required before each actor update. Table~\ref{tab:update_frequency} shows that SAG is robust across a moderate range of update frequencies. We use $K=10$ as the default because it provides a good balance between success, critic usage, and token cost.

\begin{table}[h]
\centering
\small
\setlength{\tabcolsep}{7pt}
\renewcommand{\arraystretch}{1.15}
\caption{Sensitivity to the number of successful episodes per update on ALFWorld.}
\label{tab:update_frequency}
\begin{tabular}{lccc}
\toprule
\textbf{Successes / Update} & \textbf{Succ.} $\uparrow$ & \textbf{Critic Calls / Ep.} $\downarrow$ & \textbf{\#Token} $\downarrow$ \\
\midrule
5  & 76.9 & 2.52 & 58 \\
\rowcolor{oursgray}
10 & \textbf{78.3} & \textbf{2.10} & \textbf{56} \\
20 & 78.1 & 2.28 & 57 \\
30 & 77.7 & 2.42 & 58 \\
\bottomrule
\end{tabular}
\end{table}

\subsection{Qualitative Analysis of Prompt-Induced Responses}
\label{app:critic_prompt_sensitivity}
Although the actor and critic share the same backbone models, they can produce different outputs because they operate under different prompts and objectives. This difference in prompt framing, rather than model capacity, is what allows the critic to reason beyond the actor's current step.
To examine how prompt composition shapes this behavior, we remove individual components from the full critic prompt and compare the generated rationales at identical decision points.

\begin{table}[h]
\centering
\caption{Qualitative examples of how critic prompts affect the generated rationales}
\small
\begin{tabular}{lcl}
\toprule
\textbf{Prompt variant} & \textbf{Step} & \textbf{Rationale by critic}\\
\midrule
Full prompt & 5 & Search shelves for both bowl and desklamp \\
w/o expected JSON example & 5 & Search shelves as likely bowl location \\
w/o goal & 5 &  Bowl may be on an unchecked shelf \\
\midrule
Full prompt & 9 &  Check whether bowl is near the found desklamp \\
w/o expected JSON example & 9 & Re-examine shelf for missed items \\
w/o goal & 9 &  Bowl may be near the desklamp \\
\midrule
Full prompt & 20 &  Return to desklamp to use it \\
w/o expected JSON example & 20 & Use desklamp to look at bowl \\
w/o goal & 20 & Desklamp is needed to illuminate bowl \\
\bottomrule
\end{tabular}
\label{tab:critic_prompt_sensitivity}
\end{table}

As shown in Table~\ref{tab:critic_prompt_sensitivity}, the full prompt tends to produce rationales that reflect the broader task structure, whereas the ablated prompts shift the rationale toward more local cues, such as likely object locations or immediate re-examination.
These results indicate that even with the same underlying models, the actor and critic can attend to different aspects of the task and produce meaningfully different outputs.

\subsection{Action-Only versus Reasoning-Consistent Supervision}
\label{app:sft_targets}

We compare action-only SFT with reasoning-consistent SFT,
where the critic rewrites the reasoning to support a corrected
action before supervision. Following online adaptation on 134
ALFWorld episodes, we freeze each actor and evaluate without
further updates. We report ID and OOD results, including
evaluation on 140 episode-disjoint held-out examples.

\begin{table}[htbp]
\centering
\small
\setlength{\tabcolsep}{5pt}
\caption{Comparison of SFT targets. Actor denotes execution
without critique; Calls denotes critic calls per episode.}
\label{tab:sft_targets}
\begin{tabular}{lrrrrrr}
\toprule
& \multicolumn{3}{c}{ID}
& \multicolumn{3}{c}{OOD} \\
\cmidrule(lr){2-4}\cmidrule(lr){5-7}
SFT target
& Actor SR & SAG SR & Calls
& Actor SR & SAG SR & Calls \\
\midrule
No SFT
& 24.6 & 72.3 & 3.00
& 25.7 & 70.7 & 2.85 \\
Matched reasoning
& 43.2 & 76.0 & 1.35
& 42.9 & 77.8 & 1.67 \\
Action-only
& \textbf{45.7} & \textbf{80.3} & \textbf{1.18}
& \textbf{43.4} & \textbf{80.9} & \textbf{1.41} \\
\bottomrule
\end{tabular}
\end{table}

Action-only supervision achieves higher success and fewer
critic calls than reasoning-consistent supervision in both
evaluations. These results support the effectiveness of
action-only supervision without requiring imitation of
critic-generated reasoning.

\subsection{Empirical Validation of the Gating Score}
\label{app:gate_validation}

We evaluate a frozen actor on 10 randomly sampled BabyAI
test episodes, collecting 178 non-terminal decision states.
At each actor-visited state, we record the gating score $S_t$
and whether the actor's top-1 action disagrees with BabyAI's
built-in replanning bot. We sort states by $S_t$ into four
approximately equal-sized bins, without conditioning on
whether the reference action lies among the actor's top-two
predictions.

\begin{table}[htbp]
\centering
\small
\caption{Reference-action disagreement across gating-score
quartiles, ordered from lowest to highest $S_t$.}
\label{tab:gate_validation}
\begin{tabular}{lrrrr}
\toprule
Score quartile & Q1 & Q2 & Q3 & Q4 \\
\midrule
Number of states & 45 & 45 & 44 & 44 \\
Disagreements & 8 & 10 & 14 & 20 \\
Disagreement rate (\%) & 17.8 & 22.2 & 31.8 & 45.5 \\
\bottomrule
\end{tabular}
\end{table}

Disagreement increases monotonically from 17.8\% to 45.5\%
across score quartiles, supporting the combined score as
a practical indicator of decision ambiguity. The bot is
used only as a reference for this diagnostic analysis:
disagreement is not a ground-truth error rate and does not
extend the conditional guarantee of the margin proposition.

\newpage
\section{Example Trajectories}
\label{sec:traj}

\begin{table*}[h]
\tiny
\centering
\caption{An example trajectory from ALFWorld for \texttt{pick\_two\_obj\_and\_place}. SAG selectively invokes the critic based on ambiguity (global entropy and local margin) and stores successful trajectories for online self-improvement.}
\footnotesize
\begin{tabular}{@{}l@{}}
\toprule
\begin{minipage}{0.95\linewidth}
\raggedright
\fontfamily{qcr}\selectfont
You are in the middle of a room. Looking quickly around you, you see 
an armchair 1, a cabinet 4, a cabinet 3, a cabinet 2, a cabinet 1, a drawer 5, 
a drawer 4, a drawer 3, a drawer 2, a drawer 1, a dresser 1, a garbagecan 1, 
a safe 1, a shelf 12, a shelf 11, a shelf 10, a shelf 9, a shelf 8, a shelf 7, a shelf 6, 
a shelf 5, a shelf 4, a shelf 3, a shelf 2, a shelf 1, a sidetable 1, and a sofa 1. 
Your task is to: put two pillow in sofa. 

> [Agent] Act 1: think: To solve the task, I need to find and take the first pillow, then put it in sofa, then find and take the second pillow, then put it in sofa. 

OK. 

> [Agent] Act 2: think: First I need to find the first pillow. A pillow is more likely to appear in armchair (1), cabinet (1-4), drawer (1-5), dresser (1), sidetable (1), sofa (1). I can check one by one, starting with armchair 1. 

OK. 

> [Agent] Act 3: go to armchair 1 

You arrive at armchair 1. On the armchair 1, you see a cellphone 2, a pillow 2, and a pillow 1. 

> [Agent] Act 4: take pillow 1 from armchair 1 

You pick up the pillow 1 from the armchair 1. 

> [Agent] Act 5: think: Now I find the first pillow (1). Next, I need to put it in sofa 1. 

OK. 

> [Agent] Act 6: go to sofa 1 

You arrive at sofa 1. On the sofa 1, you see a newspaper 3. 

\colorbox{yellow!20}{\parbox{\linewidth}{%
\fontfamily{qcr}\selectfont
> [Gate] global ambiguity $\bar H_t = 0.20$, local margin $m_t = 0.06 < \delta$. \\
> [Gate] score $S_t = \bar H_t + \mathbb{I}[m_t < \delta] = 1.20 > \tau$ \; (VoI proxy $>$ cost) $\Rightarrow$ critic invoked. \\[0.3em]
> [Critic] Suggested action: move \texttt{pillow 1} to \texttt{sofa 1}. \\
> [Critic] Rationale: You are close, but you forgot to put the first pillow in the sofa before searching for the second one. Putting the first pillow in the sofa now reduces the risk of forgetting it later.%
}} 

> [Agent] Act 7: move pillow 1 to sofa 1 

On the sofa 1, you see a newspaper 3, and a pillow 1. 

> [Agent] Act 8: go to armchair 1 

You arrive at armchair 1. On the armchair 1, you see a cellphone 2, and a pillow 2. 

> [Agent] Act 9: take pillow 2 from armchair 1 

You pick up the pillow 2 from the armchair 1. 

> [Agent] Act 10: think: Now I find the second pillow (2). Next, I need to put it in sofa 1. 

OK. 

> [Agent] Act 11: go to sofa 1 

You arrive at sofa 1. On the sofa 1, you see a newspaper 3, and a pillow 1. 

> [Agent] Act 12: move pillow 2 to sofa 1

You move the pillow 2 to the sofa 1. 

Outcome: success 

\colorbox{yellow!20}{\parbox{\linewidth}{%
\fontfamily{qcr}\selectfont
> [SAG] Successful episode. Trajectory is added to the self-improvement buffer.
}}

\end{minipage}\\
\bottomrule
\end{tabular}
\end{table*}

\begin{table*}[t]
\tiny
\centering
\caption{Example trajectory on ALFWorld \texttt{pick\_clean\_then\_place\_in\_recep}. SAG invokes critique when ambiguity indicates high error risk, preventing unnecessary detours.}
\footnotesize
\begin{tabular}{@{}l@{}}
\toprule
\begin{minipage}{0.95\linewidth}
\raggedright
\fontfamily{qcr}\selectfont
You are in the middle of a room. Looking quickly around you, you see 
a cabinet 4, a cabinet 3, a cabinet 2, a cabinet 1, a countertop 1, 
a garbagecan 1, a handtowelholder 2, a handtowelholder 1, 
a sinkbasin 2, a sinkbasin 1, a toilet 1, a toiletpaperhanger 1, and a towelholder 1. 
Your task is to: put a clean cloth in countertop. 

> [Agent] Act 1: think: To solve the task, I need to find and take a cloth, clean it with a sinkbasin, then place it on the countertop.  

OK.  

> [Agent] Act 2: think: First I need to find a cloth. I will check the cabinets starting from cabinet 4.  

OK.  

> [Agent] Act 3: go to cabinet 4  

You arrive at cabinet 4. The cabinet 4 is closed.  

> [Agent] Act 4: open cabinet 4  

You open the cabinet 4. The cabinet 4 is open. In it, you see a soapbottle 2.  

> [Agent] Act 5: go to cabinet 3  

You arrive at cabinet 3. The cabinet 3 is closed.  

> [Agent] Act 6: open cabinet 3  

You open the cabinet 3. The cabinet 3 is open. In it, you see a cloth 2.  

> [Agent] Act 7: take cloth 2 from cabinet 3  

You pick up the cloth 2 from the cabinet 3.  

> [Agent] Act 8: go to sinkbasin 1  

You arrive at sinkbasin 1. On the sinkbasin 1, you see a soapbar 1.  

> [Agent] Act 9: clean cloth 2 with sinkbasin 1  

You clean the cloth 2 using the sinkbasin 1.  

> [Agent] Act 10: go to countertop 1  

You arrive at countertop 1. On the countertop 1, you see a candle 1, a cloth 3, and a spraybottle 1.  

> [Agent] Act 11: think: I might need to fetch soap again before placing the cloth. I should go back to cabinet 4.  

OK.  

\colorbox{yellow!20}{\parbox{\linewidth}{%
\fontfamily{qcr}\selectfont
> [Gate] global ambiguity $\bar H_t = 0.55$, local margin $m_t = 0.78$. \\
> [Gate] score $S_t = \bar H_t + \mathbb{I}[m_t < \delta] = 0.55 > \tau$ \; (VoI proxy $>$ cost) $\Rightarrow$ critic invoked.
> [Critic] Suggested action: move \texttt{cloth 2} to \texttt{countertop 1}. \\
> [Critic] Rationale: The cloth you are holding is already clean. The task only requires placing a clean cloth on the countertop. Fetching soap again is unnecessary and risks delaying completion.%
}} 

> [Agent] Act 12: move cloth 2 to countertop 1

On the countertop 1, you see a candle 1, a cloth 3, and a spraybottle 1.  

> [Agent] Act 13: move cloth 2 to countertop 1  

You move the cloth 2 to the countertop 1.  

Outcome: success

\colorbox{yellow!20}{\parbox{\linewidth}{%
\fontfamily{qcr}\selectfont
> [SAG] Successful episode. The trajectory is added to the self-improvement buffer $\mathcal{D}$.
}}

\end{minipage}\\
\bottomrule
\end{tabular}
\end{table*}

\begin{table*}[t]
\tiny
\centering
\caption{Example trajectory on ALFWorld \texttt{pick\_heat\_then\_place\_in\_recep}. SAG corrects inefficient exploration by invoking critique under high global ambiguity.}
\footnotesize
\begin{tabular}{@{}l@{}}
\toprule
\begin{minipage}{0.95\linewidth}
\raggedright
\fontfamily{qcr}\selectfont
You are in the middle of a room. Looking quickly around you, you see 
a cabinet 6, a cabinet 5, a cabinet 4, a cabinet 3, a cabinet 2, a cabinet 1, 
a coffeemachine 1, a countertop 3, a countertop 2, a countertop 1, 
a drawer 3, a drawer 2, a drawer 1, a fridge 1, a garbagecan 1, 
a microwave 1, a shelf 3, a shelf 2, a shelf 1, a sinkbasin 1, 
a stoveburner 4, a stoveburner 3, a stoveburner 2, a stoveburner 1, 
and a toaster 1.  \\
Your task is to: put a hot apple in fridge.  

> [Agent] Act 1: think: First I need to find an apple. I will start by checking the fridge.  

OK.  

> [Agent] Act 2: go to fridge 1  

You arrive at fridge 1. The fridge 1 is closed.  

> [Agent] Act 3: open fridge 1  

You open the fridge 1. In it, you see a bowl 1, a cup 2, a cup 1, a lettuce 1, a mug 2, a potato 1, and a tomato 1.  

> [Agent] Act 4: think: No apple here. I might need to search cabinets one by one.  

OK.  

\colorbox{yellow!20}{\parbox{\linewidth}{%
\fontfamily{qcr}\selectfont
> [Gate] global ambiguity $\bar H_t = 0.21$, local margin $m_t = 0.09 < \delta$. \\
> [Gate] score $S_t = \bar H_t + \mathbb{I}[m_t < \delta] = 1.21 > \tau$ \; (VoI proxy $>$ cost) $\Rightarrow$ critic invoked. \\[0.3em]
> [Critic] Suggested action: go to \texttt{countertop 1}. \\
> [Critic] Rationale: Apples are more likely to be placed on open surfaces such as countertops. Searching cabinets first is less efficient.
}}  

> [Agent] Act 5: go to countertop 1  

You arrive at countertop 1. On the countertop 1, you see an apple 1, an apple 2, and a mug 3.  

> [Agent] Act 6: take apple 1 from countertop 1  

You pick up the apple 1 from the countertop 1.  

> [Agent] Act 7: go to microwave 1  

You arrive at microwave 1. The microwave 1 is closed.  

> [Agent] Act 8: heat apple 1 with microwave 1  

You heat the apple 1 using the microwave 1.  

> [Agent] Act 9: think: The apple is now hot. I should place it in the fridge.  

OK.  

> [Agent] Act 10: go to fridge 1  

You arrive at fridge 1. The fridge 1 is open.  

> [Agent] Act 11: move apple 1 to fridge 1  

You move the apple 1 to the fridge 1.  

Outcome: success 

\colorbox{yellow!20}{\parbox{\linewidth}{%
\fontfamily{qcr}\selectfont
> [SAG] Successful trajectory is added to the self-improvement buffer $\mathcal{D}$.
}}

\end{minipage}\\
\bottomrule
\end{tabular}
\end{table*}

\begin{table*}[h]
\tiny
\centering
\caption{Example trajectory on BabyAI-Text \texttt{pick\_up\_green\_key}. SAG intervenes selectively to prevent looping behavior.}
\footnotesize
\begin{tabular}{@{}l@{}}
\toprule
\begin{minipage}{0.95\linewidth}
\raggedright
\fontfamily{qcr}\selectfont
Your mission: pick up the green key. You start in a corridor with a blue key on the floor and a closed door in front.

> [Agent] Act 1: think: The green key is not visible. I should move toward the door.  
OK.  

> [Agent] Act 2: go forward  

You are now standing directly in front of the door.

\colorbox{yellow!20}{\parbox{\linewidth}{%
\fontfamily{qcr}\selectfont
> [Gate] global ambiguity $\bar H_t = 0.28$, local margin $m_t = 0.08 < \delta$. \\
> [Gate] score $S_t = 1.28 > \tau \Rightarrow$ critic invoked. \\[0.3em]
> [Critic] Suggested action: open the door. \\
> [Critic] Rationale: Progress requires entering the next room; waiting in front of the door adds no information.
}}

> [Agent] Act 3: open door  

The door opens. Inside you see a red ball in front, a green key on the right, and a yellow box.  

> [Agent] Act 4: go forward  

> [Agent] Act 5: turn left  

> [Agent] Act 6: turn right  

\colorbox{yellow!20}{\parbox{\linewidth}{%
\fontfamily{qcr}\selectfont
> [Gate] global ambiguity $\bar H_t = 0.34$, local margin $m_t = 0.07 < \delta$. \\
> [Gate] score $S_t = 1.34 > \tau \Rightarrow$ critic invoked. \\[0.3em]
> [Critic] Suggested action: turn right toward the green key. \\
> [Critic] Rationale: The goal object is already identified; reorientation loops are unnecessary.
}}  

> [Agent] Act 7: turn right  

> [Agent] Act 8: go forward  

> [Agent] Act 9: pick up green key  

> [Agent] Act 10: stop  

Outcome: success  

\colorbox{yellow!20}{\parbox{\linewidth}{%
\fontfamily{qcr}\selectfont
> [SAG] Successful trajectory with two gated interventions is added to the self-improvement buffer $\mathcal{D}$.%
}}  
\end{minipage}\\
\bottomrule
\end{tabular}
\end{table*}

\begin{table*}[h]
\tiny
\centering
\caption{Example trajectory on WebShop \texttt{buy\_red\_tshirt\_under\_20}. SAG resolves constraint ambiguity via targeted critique.}
\footnotesize
\begin{tabular}{@{}l@{}}
\toprule
\begin{minipage}{0.95\linewidth}
\raggedright
\fontfamily{qcr}\selectfont
Your task: purchase a red T-shirt under \$20. The homepage lists categories: shirts, shoes, bags.

> [Agent] Act 1: click "shirts"  

The page shows 5 items: blue T-shirt (\$15), red T-shirt (\$25), red T-shirt (\$18), green hoodie (\$22), white T-shirt (\$10).

> [Agent] Act 2: think: There are two red T-shirts. I might choose the \$25 one.  

\colorbox{yellow!20}{\parbox{\linewidth}{%
\fontfamily{qcr}\selectfont
> [Gate] global ambiguity $\bar H_t = 0.33$, local margin $m_t = 0.04 < \delta$. \\
> [Gate] score $S_t = 1.33 > \tau \Rightarrow$ critic invoked. \\[0.3em]
> [Critic] Suggested action: add the \$18 red T-shirt to the cart. \\
> [Critic] Rationale: The \$25 option violates the budget constraint.
}}  

> [Agent] Act 3: add red T-shirt (\$18) to cart  

> [Agent] Act 4: buy

Outcome: Reward 1.0, success  

\colorbox{yellow!20}{\parbox{\linewidth}{%
\fontfamily{qcr}\selectfont
> [SAG] Successful trajectory is added to self-improvement buffer.%
}}  
\end{minipage}\\
\bottomrule
\end{tabular}
\end{table*}

\begin{table*}[h]
\tiny
\centering
\caption{Example trajectory on WebShop \texttt{buy\_black\_backpack}. SAG intervenes twice to enforce constraints and finalize execution.}
\footnotesize
\begin{tabular}{@{}l@{}}
\toprule
\begin{minipage}{0.95\linewidth}
\raggedright
\fontfamily{qcr}\selectfont
Your task: purchase a cheap black backpack.

> [Agent] Act 1: click "bags"  

The bags page shows: black backpack (\$55), black backpack (\$25), red backpack (\$20).

> [Agent] Act 2: think: The \$55 one might be better quality.  

\colorbox{yellow!20}{\parbox{\linewidth}{%
\fontfamily{qcr}\selectfont
> [Gate] global ambiguity $\bar H_t = 0.24$, local margin $m_t = 0.07 < \delta$. \\
> [Gate] score $S_t = 1.24 > \tau \Rightarrow$ critic invoked. \\[0.3em]
> [Critic] Suggested action: add the \$25 black backpack to the cart. \\
> [Critic] Rationale: The task requires a cheap black backpack; \$55 violates this constraint.
}}  

> [Agent] Act 3: add black backpack (\$25) to cart  

> [Agent] Act 4: view cart  

> [Agent] Act 5: think: The cart satisfies color and price.  

> [Agent] Act 6: go to checkout  

\colorbox{yellow!20}{\parbox{\linewidth}{%
\fontfamily{qcr}\selectfont
> [Gate] global ambiguity $\bar H_t = 0.31$, local margin $m_t = 0.06 < \delta$. \\
> [Gate] score $S_t = 1.31 > \tau \Rightarrow$ critic invoked. \\[0.3em]
> [Critic] Suggested action: place the order. \\
> [Critic] Rationale: No further browsing is required once all constraints are satisfied.
}}  

> [Agent] Act 7: buy  

Outcome: Reward 1.0, success  

\end{minipage}\\
\bottomrule
\end{tabular}
\end{table*}

\newpage
\clearpage
\section{GenAI Usage Disclosure}
\label{app: genai}

We acknowledge the use of large language model (LLM) assistants in the preparation of this paper as follows:

\begin{itemize}
    \item \textbf{Writing assistance:} ChatGPT (OpenAI) was used to improve grammar, rephrase sentences, and refine the clarity of certain passages. All conceptual content, experimental design, analysis, and claims were authored solely by the listed authors.
    \item \textbf{Code generation:} No LLM-based tools were used for writing or generating code in this study.
    \item \textbf{Data processing or analysis:} No LLM-based tools were used for data preprocessing, analysis, or the generation of results.
\end{itemize}

\newpage
\clearpage
\section*{NeurIPS Paper Checklist}

\begin{enumerate}
\item {\bf Claims}
    \item[] Question: Do the main claims made in the abstract and introduction accurately reflect the paper's contributions and scope?
    \item[] Answer: \answerYes{}
    \item[] Justification:  We have claimed our scope, and contribution in the section of Introduction.

\item {\bf Limitations}
    \item[] Question: Does the paper discuss the limitations of the work performed by the authors?
    \item[] Answer: \answerYes{} 
    \item[] Justification: We discuss the limitations in Appendix~\ref{Impact}.

\item {\bf Theory assumptions and proofs}
    \item[] Question: For each theoretical result, does the paper provide the full set of assumptions and a complete (and correct) proof?
    \item[] Answer: \answerYes{} 
    \item[] Justification: The assumptions and theoretical results are provided in Section~\ref{sec:theory_proxy}. Complete mathematical proofs are provided in Appendix~\ref{app:voi_gate}, ~\ref{app:entropy_bounds}, and ~\ref{app:margin_risk}.

    \item {\bf Experimental result reproducibility}
    \item[] Question: Does the paper fully disclose all the information needed to reproduce the main experimental results of the paper to the extent that it affects the main claims and/or conclusions of the paper (regardless of whether the code and data are provided or not)?
    \item[] Answer: \answerYes{} 
    \item[] Justification: We provide the main experimental settings and hyperparameters needed to reproduce the results in Section~\ref{experimental_setting} and Appendix~\ref{app:prompt} and ~\ref{app:setting}.
    

\item {\bf Open access to data and code}
    \item[] Question: Does the paper provide open access to the data and code, with sufficient instructions to faithfully reproduce the main experimental results, as described in supplemental material?
    \item[] Answer: \answerYes{} 
    \item[] Justification: We provide the source code for our proposed method and scripts to reproduce the experimental results.

\item {\bf Experimental setting/details}
    \item[] Question: Does the paper specify all the training and test details (e.g., data splits, hyperparameters, how they were chosen, type of optimizer) necessary to understand the results?
    \item[] Answer: \answerYes{} 
    \item[] Justification: Dataset descriptions and experimental settings are provided in Section~\ref{experimental_setting}, and detailed hyperparameters are provided in Appendix~\ref{app:prompt} and ~\ref{app:setting}.

\item {\bf Experiment statistical significance}
    \item[] Question: Does the paper report error bars suitably and correctly defined or other appropriate information about the statistical significance of the experiments?
    \item[] Answer: \answerYes{} 
    \item[] Justification: We report the mean and one standard deviation to indicate the reliability of the results, as shown in Appendix ~\ref{app: shuffled_order}.

\item {\bf Experiments compute resources}
    \item[] Question: For each experiment, does the paper provide sufficient information on the computer resources (type of compute workers, memory, time of execution) needed to reproduce the experiments?
    \item[] Answer: \answerYes{} 
    \item[] Justification: Computational resources are reported in Appendix~\ref{app:hyperparameter}.
    
\item {\bf Code of ethics}
    \item[] Question: Does the research conducted in the paper conform, in every respect, with the NeurIPS Code of Ethics \url{https://neurips.cc/public/EthicsGuidelines}?
    \item[] Answer: \answerYes{} 
    \item[] Justification: The research conforms to the NeurIPS Code of Ethics.

\item {\bf Broader impacts}
    \item[] Question: Does the paper discuss both potential positive societal impacts and negative societal impacts of the work performed?
    \item[] Answer: \answerYes{} 
    \item[] Justification: Broader impacts are reported in Appendix~\ref{Impact}.
    
\item {\bf Safeguards}
    \item[] Question: Does the paper describe safeguards that have been put in place for responsible release of data or models that have a high risk for misuse (e.g., pre-trained language models, image generators, or scraped datasets)?
    \item[] Answer: \answerNA{} 
    \item[] Justification: \answerNA{}

\item {\bf Licenses for existing assets}
    \item[] Question: Are the creators or original owners of assets (e.g., code, data, models), used in the paper, properly credited and are the license and terms of use explicitly mentioned and properly respected?
    \item[] Answer: \answerYes{} 
    \item[] Justification: In our experiments, we used datasets and models that are all open-source and widely utilized. 

\item {\bf New assets}
    \item[] Question: Are new assets introduced in the paper well documented and is the documentation provided alongside the assets?
    \item[] Answer: \answerYes{} 
    \item[] Justification: We provide the implementation with documentation and reproduction instructions.

\item {\bf Crowdsourcing and research with human subjects}
    \item[] Question: For crowdsourcing experiments and research with human subjects, does the paper include the full text of instructions given to participants and screenshots, if applicable, as well as details about compensation (if any)? 
    \item[] Answer: \answerNA{} 
    \item[] Justification: \answerNA{}

\item {\bf Institutional review board (IRB) approvals or equivalent for research with human subjects}
    \item[] Question: Does the paper describe potential risks incurred by study participants, whether such risks were disclosed to the subjects, and whether Institutional Review Board (IRB) approvals (or an equivalent approval/review based on the requirements of your country or institution) were obtained?
    \item[] Answer: \answerNA{} 
    \item[] Justification: \answerNA{}

\item {\bf Declaration of LLM usage}
    \item[] Question: Does the paper describe the usage of LLMs if it is an important, original, or non-standard component of the core methods in this research? Note that if the LLM is used only for writing, editing, or formatting purposes and does \emph{not} impact the core methodology, scientific rigor, or originality of the research, declaration is not required.
    \item[] Answer: \answerNA{} 
    \item[] Justification: LLMs were used only for writing assistance, which does not affect the core methodology. We disclose this in Appendix~\ref{app: genai}.

\end{enumerate}

\end{document}